\documentclass{article} 
\usepackage{iclr2021_conference,times}

\usepackage{amsmath,amsfonts,bm}

\def\eqref#1{equation~\ref{#1}}

\def\1{\bm{1}}

\def\rx{{\textnormal{x}}}
\def\ry{{\textnormal{y}}}
\def\rz{{\textnormal{z}}}

\def\rvepsilon{{\mathbf{\epsilon}}}

\def\rvx{{\mathbf{x}}}
\def\rvy{{\mathbf{y}}}
\def\rvz{{\mathbf{z}}}

\def\rmA{{\mathbf{A}}}

\def\rmX{{\mathbf{X}}}

\def\rmZ{{\mathbf{Z}}}

\def\vx{{\bm{x}}}
\def\vy{{\bm{y}}}

\def\mA{{\bm{A}}}

\def\mD{{\bm{D}}}

\def\mI{{\bm{I}}}

\def\mT{{\bm{T}}}
\def\mU{{\bm{U}}}

\def\mBeta{{\bm{\beta}}}

\def\mLambda{{\bm{\Lambda}}}
\def\mSigma{{\bm{\Sigma}}}
\def\mPi{{\bm{\Pi}}}

\DeclareMathAlphabet{\mathsfit}{\encodingdefault}{\sfdefault}{m}{sl}
\SetMathAlphabet{\mathsfit}{bold}{\encodingdefault}{\sfdefault}{bx}{n}

\def\sC{{\mathbb{C}}}

\def\sR{{\mathbb{R}}}

\newcommand{\E}{\mathbb{E}}

\newcommand{\Var}{\mathrm{Var}}

\newcommand{\Cov}{\mathrm{Cov}}

\DeclareMathOperator*{\argmin}{arg\,min}

\DeclareMathOperator{\Tr}{Tr}

\usepackage{amsmath,epsfig,amssymb,amsfonts,amsthm,verbatim,epstopdf,float}
\usepackage{hyperref}
\usepackage{url}
\usepackage{bm}
\usepackage{dsfont}
\usepackage{esint}

\title{Provable More Data Hurt in High Dimensional Least Squares Estimator}

\iclrfinalcopy

\author{Zeng Li \\
Department of Statistics and Data Science \\
Southern University of Science and Technology \\
Shenzhen, China \\
\texttt{liz9@sustech.edu.cn} \\
\And
Chuanlong Xie \\ 
Huawei Noah's Ark Lab\\
Hong Kong, China \\
\texttt{xie.chuanlong@huawei.com} \\
\And
Qinwen Wang \thanks{corresponding author} \\
School of Data Science \\
Fudan University\\
Shanghai, China \\
\texttt{wqw@fudan.edu.cn}
}

\newtheorem{theorem}{Theorem}[section]

\newtheorem{lemma}{Lemma}[section]
\newtheorem{remark}{Remark}[section]

\newcommand{\T}{{\mathrm{\scriptscriptstyle T}}}
\newcommand{\rarrow}{\rightarrow}

\newcommand{\benr}{\begin{eqnarray}}
\newcommand{\eenr}{\end{eqnarray}}
\newcommand{\benrr}{\begin{eqnarray*}}
\newcommand{\eenrr}{\end{eqnarray*}}
\newcommand{\ben}{\begin{equation}}
\newcommand{\een}{\end{equation}}
\newcommand{\benn}{\begin{equation*}}
\newcommand{\eenn}{\end{equation*}}

\newcommand{\tecr}{\textcolor{red}}

\newcommand{\bbox}{\hfill $\Box$}

\newcommand\red[1]{{\color{black}#1}}

\begin{document}

\maketitle

\begin{abstract}
This paper investigates the finite-sample prediction risk of the high-dimensional least squares estimator.
We derive the central limit theorem for the prediction risk when both the sample size and the number of features tend to infinity.
Furthermore, the finite-sample distribution and the confidence interval of the prediction risk are provided.
Our theoretical results demonstrate the sample-wise non-monotonicity of the prediction risk and confirm ``more data hurt'' phenomenon.
\end{abstract}

\section{Introduction}

More data hurt refers to the phenomenon that training on more data can hurt the prediction performance of the learned model, especially for some deep learning tasks.
\red{\cite{loog2019minimizers} shows that various standard learners can lead to sample-wise non-monotonicity in linear model.}
\cite{nakkiran2019deep} experimentally confirms the
sample-wise non-monotonicity of the test accuracy on deep neural networks.
This challenges the conventional understanding in large sample properties: if an estimator is  consistent, more data makes the estimator more stable and improves its finite-sample performance. 
\cite{nakkiran2019more} considers adding one single data point to a linear regression task and analyzes its marginal effect to the test risk. 
\cite{derezinski2019exact} gives an exact non-asymptotic risk of the high-dimensional least squares estimator, and observes the sample-wise non-monotonicity on MSE. 
For adversarially robust models,
\red{\cite{min2020curious} proves that more
data may increase the gap between the generalization
error of adversarially-trained models and standard models. 
\cite{chen2020more} shows that more training data causes the generalization error to increase in the strong adversary regime.}
In this work, we derive the finite-sample distribution of the prediction risk \red{under linear model} and prove  the \emph{``more data hurt"} phenomenon from asymptotic point of view.

Intuitively, the \emph{``more data hurt"} stems from the \emph{``double descent"} risk curve: as the model complexity increases, the prediction risk of the learned model
first decreases and then increases, and then decreases again. 
The \emph{double descent} phenomenon can be precisely quantified for certain simple models (\cite{hastie2019surprises,mei2019generalization,ba2019generalization,belkin2019two,bartlett2020benign,xing2019benefit}).
Among these works, \cite{hastie2019surprises} and \cite{mei2019generalization} use the tools from random matrix theory and explicitly prove the double descent curve of the asymptotic risk of linear regression and
random features regression in high dimensional setup. 
\cite{ba2019generalization} gives the
asymptotic risk of two-layer neural networks when either the first or the second layer is trained using a gradient flow.

The second decline of the prediction risk in the double descent curve is highly related to the \emph{more data hurt} phenomenon.
In the over-parameterized regime when the model complexity is fixed while the sample size increases, the degree of over-parameterization decreases and becomes close to the interpolation boundary (for example $p/n=1$ in \cite{hastie2019surprises}), in which a high prediction risk is achieved.
However, the existing asymptotic results, which focus on the first order limit of the prediction risk, cannot exactly guarantee the \emph{more data hurt} phenomenon. Hence, in this work, we characterize the second order fluctuations of the prediction risk and make attempts to fill this gap.
We employ the linear regression task in \cite{hastie2019surprises} and \cite{nakkiran2019more}, and introduce new tools from the random matrix theory, e.g. the central limit theorem for linear spectral statistics in \cite{bai2004clt, Bai2007}, to derive the central limit theorem of the prediction risk.

Consider a linear regression task with $n$ data points and $p$ features, 
the setup of the \emph{more data hurt} is similar with that in the classical asymptotic analysis in \cite{van2000asymptotic}.
According to the classical asymptotic analysis with $p$ fixed and $n\rarrow \infty$, the least square estimator is unbiased and $\sqrt n$-consistent to the ground truth. 
This implies that the more data will not hurt and even improve the prediction performance when $p<n$ and the sample size is sufficiently large. However, the story is very different in the overparameterized regime. The prediction risk doesn't decrease monotonously with $n$ when $p>n$. More data does hurt in the overparametrized case.
In the following, we will justify this phenomenon by developing the CLT results as both $n$ and $p$ tend to infinity. 
We assume $p/n \rarrow c$, and denote $0<n_1<n_2<+\infty$, $c_1 = p/n_1$ and $c_2=p/n_2.$ 
Then the direct comparison of the prediction risk between sample sizes  
$n_1$ and $n_2$ can be decomposed into three parts: (i) the gap between the finite-sample risk under $n=n_1$ and the asymptotic risk with $c=c_1$; (ii) the gap between the finite-sample risk under $n=n_2$ and the asymptotic risk with $c=c_2$;  (iii) the comparison between two asymptotic risk under $c=c_1$ and $c=c_2$.
Theorem~1 and~2 of \cite{hastie2019surprises} give answers to task (iii). 
For (i) and (ii), we develop the convergence rate and the limiting distribution of the prediction risk as $n,p\rarrow +\infty$, $p/n\rarrow c$ in this paper.
Thus the finite-sample distribution of the prediction risk can be approximated by its limiting distribution.
Furthermore, the confidence interval of the finite-sample risk can be obtained as well.

\begin{figure}[ht]
\begin{center}
\includegraphics[width=\columnwidth]{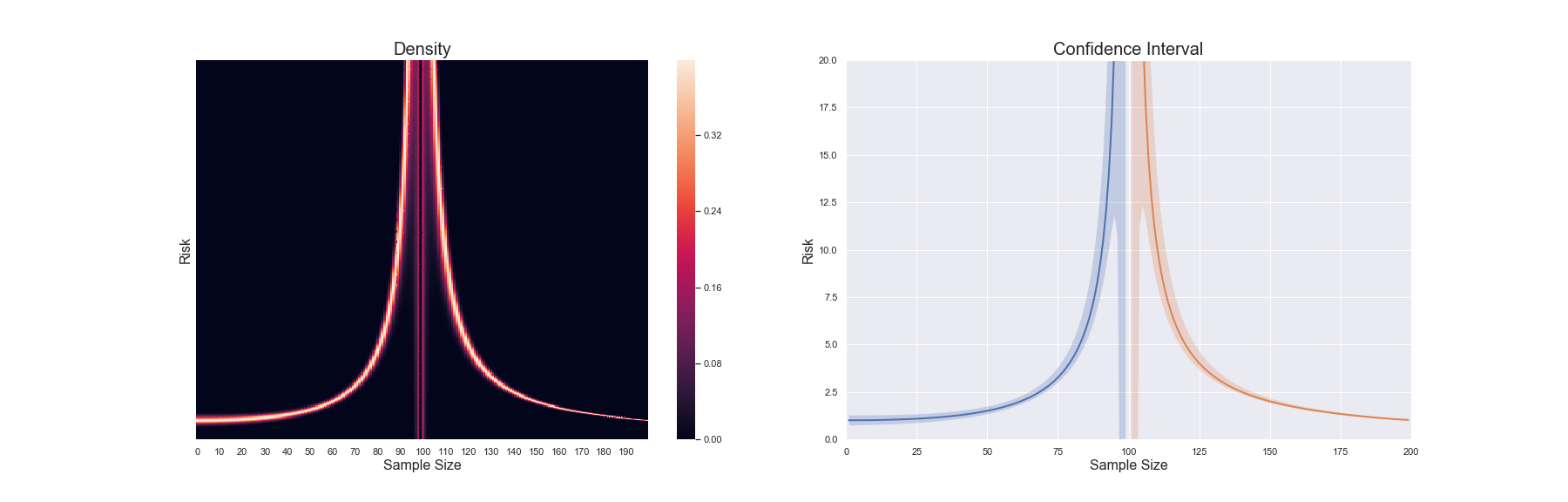}
\end{center}
\caption{Sample-wise double descent.  
We take $p=100$ and $1\leq n\leq 200.$ {\bf Left}: The conditional density of the prediction risk when sample size varies from 1 to 200.
According to the conditional distribution of the prediction risk, we can clearly observe the 
 sample-wise double descent phenomenon. {\bf Right}: The 95$\%$-confidence band (point-wise) of the prediction risk. In the overparameterized regime $1\leq n < 100$, there exists some pairs $(n_1, n_2)$, $1\leq n_1 < n_2 < 100$ such that the upper boundary of the confidence interval at $n_1$ is smaller than the lower boundary of the confidence interval at $n_2.$ This confirms the more data hurt phenomenon.}
\label{FigIntro}
\end{figure}

We summarize our findings as follows:
\begin{itemize}
\item The finite-sample distribution of the prediction risk is derived and the sample-wise double descent is characterized in Theorem~\ref{thm:underparaRXB} and Theorem~\ref{thm:overparaRXB} (see Figure~\ref{FigIntro}). Under certain assumptions, the more data hurt phenomenon can be confirmed by comparing the confidence intervals built via the central limit theorems. 
\item Two different types of prediction risk in the linear regression model are considered in Section~\ref{sec4}, one conditional risk given both the training data and regression coefficient, the other conditional risk given the training data only. The regression coefficient is set to be either random or nonrandom to cover more cases. Different convergence rates and limiting distributions of both prediction risk are derived under various scenarios.
\item Our results incorporate non-Gaussian observations. 
For Gaussian data, the limiting mean and variance in the central limit theorems have simpler forms, see Section~\ref{sec42} and \ref{sec43} for more details.
\end{itemize}

\section{Related work}

{\bf Double Descent}
The \emph{double descent} curve describes how generalization ability changes as model capacity increases.
It subsumes the classical bias-variance trade-off, a U-shape curve, and further show that the test error exhibits a second drop when the model capacity exceeds the interpolation threshold (\cite{belkin2018reconciling,geiger2019jamming,spigler2019jamming,advani2017high}).
The \emph{double descent} phenomenon has been quantified for certain models, including two layer neural networks via non-asymptotic bounds or asymptotic risk (\cite{belkin2019two,muthukumar2020harmless,hastie2019surprises,mei2019generalization,ba2019generalization}).
As our results are based on linear regression model, we focus on the literature of linear models.
\cite{muthukumar2020harmless} and \cite{bartlett2020benign} derive the generalization bounds for overparametrized linear models and show the benefits of the interpolation.
\cite{hastie2019surprises} gives the first order limit of
the generalization error for linear regressions as $n,p\rarrow +\infty.$
\cite{derezinski2019exact} provides an exact non-asymptotic expressions for \emph{double descent} of the high-dimensional least square estimator.
\cite{montanari2019generalization}, \cite{deng2019model} and \cite{kini2020analytic} investigate the shape asymptotics of binary classification tasks with the max-margin solution and the maximum likelihood solution.
\cite{emami2020generalization} and \cite{gerbelot2020asymptotic} consider the \emph{double descent} in generalized linear models.
Furthermore, the \emph{double descent} phenomenon is also observed on  linear tasks with various problems and assumptions, e.g. \cite{lejeune2020implicit,gerbelot2020asymptoticb,javanmard2020precise,dar2020double,xu2019number,dar2020subspace}.  
\cite{xing2019benefit} sharply quantifies the benefit
of interpolation in the nearest neighbors algorithm.
\cite{mei2019generalization} derives the limit risk on the random features model, and shows that minimum generalization error is achieved by highly overparametrized interpolators.
\cite{ba2019generalization} gives the
limit risk of the regression problem
under two-layer neural networks.
However, the existing asymptotic results focus on the first order limit of prediction risk and do not indicate the convergence rate. 
In this work, we are the first to develop results on  second order fluctuations of the prediction risk in linear regressions and provide its corresponding confidence intervals. The more data hurt phenomenon is further justified from the asymptotic point of view.

{\bf Random Matrix Theory} The primary tool for analyzing the second order fluctuations of prediction risk comes from random matrix theory. In particular, \cite{bai2004clt} refines the central limit theorem for linear spectral statistics of large dimensional sample covariance matrix with general population and the population is not necessary to be Gaussian. Such central limit theorems are also developed for other random matrix ensembles, see \cite{sinai1998central, BY05, zheng12}.
Other than the central limit theorem for linear spectral statistics, \cite{Bai2007} and \cite{pan2008}  study the asymptotic fluctuation of eigenvectors of sample covariance matrices.   \cite{BY08} considers quadratic forms like the type $\rvx_0^\T\rmA\rvx_0$. 
All these technical tools and results are adopted and fully utilized in this paper, especially those based on Stieltjes transform that are closely related to the prediction risk studied in this paper.

The main goal of this paper is to study the asymptotic behavior of two different types of prediction risk in the linear regression model. The rest of this paper is organized as follows. Section~\ref{sec:pre} introduces the model settings and 
two different prediction risk. Section~\ref{sec4} presents the main results on CLTs for the two types of risk. Section~\ref{sec:simu} conducts simulation experiments to verify the main results. All the technical proofs and lemmas are relegated to the appendix in the supplementary file.

\section{Preliminaries}\label{sec:pre}

\subsection{Problem, data and estimator}\label{sec31}

Suppose that the training data $\{ (\rvx_i, \ry_i)\in \sR^{p} \times \sR, i=1, 2, \ldots, n\}$ is generated from the model (ground truth or teacher model):
\benr\label{eq11}
\ry_i =  \mBeta^\T \rvx_i + \rvepsilon_i, \quad \text{and} \quad (\rvx_i, \rvepsilon_i)\sim (P_{\rvx}, P_\rvepsilon), \quad  i=1,2, \ldots, n,
\eenr
where the randomness across $i=1, \ldots, n$ is independent. 
Here, $P_\rvx$ is a distribution on $\sR^p$ such that $\E(\rvx_i) = {\bf 0}$, $\Cov(\rvx_i) = \mSigma$, and $P_\rvepsilon$ is a distribution on $\sR$ such that $\E(\rvepsilon_i) = 0$, $\Var(\rvepsilon_i) = \sigma^2.$ 
To proceed further, we denote
\benrr
\rmX_{n\times p} = (\rvx_1, \rvx_2, \ldots, \rvx_n)^\T, \quad \rvy = (\ry_1, \ry_2, \ldots, \ry_n)^\T.
\eenrr
The minimum $\ell_2$ norm (min-norm) least squares estimator, of $\rvy$ on $\rmX$, is defined by
\benr\label{eq12}
\hat\mBeta = \argmin_\mBeta \|\rvy-\rmX\mBeta \|^2 =  (\rmX^\T \rmX)^{+} \rmX^\T \rvy,
\eenr
where $(\rmX^\T \rmX)^{+}$ denotes the Moore-Penrose pseudoinverse of $\rmX^\T \rmX.$


\subsection{Bias, variance and risk}\label{sec32}

Similar to \cite{hastie2019surprises}, we define two different types of out-of-sample prediction risk. 
The first one is  given by
\benrr
R_\rmX (\hat\mBeta, \mBeta) = \E\big[(\rvx_0^\T \hat\mBeta - \rvx_0^\T \mBeta)^2\big|\rmX\big] 
= \E\big[ \|\hat\mBeta-\mBeta\|_{\mSigma}^2 \big| \rmX \big],
\eenrr
where $\rvx_0 \sim P_\rvx$ is a test point and is independent of the training data, and $\|\mBeta\|_{\mSigma}^2$ stands for $\mBeta^\T \mSigma \mBeta.$
Here $\mBeta$ is assumed to be a random vector independent of $\rvx_0.$
In this definition, the expectation $\E$ stands for the conditional expectation with respect to $\rvx_0$, $\hat\mBeta$ and $\mBeta$ when $\rmX$ is given.
According to the bias-variance decomposition, we have
$R_\rmX (\hat\mBeta, \mBeta): = B_\rmX (\hat\mBeta, \mBeta) + V_\rmX (\hat\mBeta, \mBeta)$, where
\ben\label{eq13}
B_\rmX (\hat\mBeta, \mBeta) = \E\Big\{\|\E(\hat\mBeta|\rmX) - \mBeta\|_{\mSigma}^2\big| \rmX \Big\} \quad \text{and} \quad V_\rmX (\hat\mBeta, \mBeta) = \Tr\{\Cov(\hat\mBeta|\rmX)\mSigma\}. 
\een
Plugging the model (\ref{eq11}) into the min-norm estimator (\ref{eq12}), the bias and variance terms can be rewritten as
\benrr
B_\rmX (\hat\mBeta, \mBeta) = \E\big\{\mBeta^\T \Pi \mSigma \Pi \mBeta\big| \rmX \big\} \quad \text{and} \quad V_\rmX (\hat\mBeta, \mBeta) = \frac{\sigma^2}{n} \Tr(\hat\mSigma^{+}\mSigma),
\eenrr
where $\hat \mSigma = \rmX^\T \rmX/n$ is the (uncentered) sample covariance matrix of $\rmX$, and $\mPi = \mI_p -  \hat\mSigma^{+}\hat\mSigma$ is the projection onto the null space of $\rmX.$
	
The second type of out-of-sample prediction risk is defined as
\benn
R_{\rmX,\mBeta} (\hat\mBeta, \mBeta) = \E\big[(\rvx_0^\T \hat\mBeta - \rvx_0^\T \mBeta)^2\big|\rmX,\mBeta\big]= \E\big[ \|\hat\mBeta-\mBeta\|_{\mSigma}^2 \big| \rmX,\mBeta \big],
\eenn
where
\benn
B_{\rmX,\mBeta} (\hat\mBeta, \mBeta)  = \mBeta^\T \Pi \mSigma \Pi \mBeta \quad \text{and} \quad V_{\rmX,\mBeta} (\hat\mBeta, \mBeta) = V_\rmX (\hat\mBeta, \mBeta) = \frac{\sigma^2}{n} \Tr(\hat\mSigma^{+}\mSigma).
\eenn
In this definition, the parameter $\mBeta$ is assumed to be given. The expectation $\E$ is the conditional expectation with respect to $\rvx_0$ and $\hat\mBeta$  when $\rmX$ and $\mBeta$ are given.
This is consistent with the common-used testing procedure, in which a trained model is evaluated by the average loss on unseen testing data. Our main goal is to study the asymptotic behavior of the two types of out-of-sample prediction risk $R_{\rmX}$ and $R_{\rmX,\mBeta}$ as $n, p \rarrow +\infty$ and $p/n \rarrow c \in (0, +\infty).$


\section{Main Results}\label{sec4}

Before stating our main results, we briefly highlight the challenges we faced in proving the \emph{more data hurt} phenomenon. First, the finite-sample behavior of prediction risk is required. \cite{hastie2019surprises} gives the first order limit of both $R_{\rmX,\mBeta} (\hat\mBeta, \mBeta)$ and $R_{\rmX} (\hat\mBeta, \mBeta)$
as $n, p \rarrow +\infty$ and $p/n \rarrow c \in (0, +\infty).$
However, to prove the \emph{more data hurt} phenomenon, we should fix $p$ and investigate the finite-sample risk with different sample sizes $n$.
This implies that only knowing the first order limit is not enough,  the convergence rate is also needed.
To solve this problem, we have derived the central limit theorems for $R_{\rmX,\mBeta} (\hat\mBeta, \mBeta)$ and $R_{\rmX} (\hat\mBeta, \mBeta)$ respectively, which characterize the second order fluctuations of the risk.
Then we can figure out the finite-sample behavior of the risk by computing the gap between the risk and its limit. 
The confidence intervals of the risk can be further obtained. 
Second, the parameter $\mBeta$ also contributes randomness to the finite-sample risk, which further influences the convergence rate. To analyze the contribution of $\mBeta$, we need to make use of the technical tools and asymptotic results for eigenvectors and quadratic forms developed in \cite{Bai2007} and \cite{BY08}.
Another interesting finding is that, in  the overparameterized regime
such that $p>n$, the two types of out-of-sample prediction risk
$R_{\rmX,\mBeta} (\hat\mBeta, \mBeta)$ and $R_{\rmX} (\hat\mBeta, \mBeta)$ actually
 enjoy
different convergence rates.

\subsection{Assumptions and more notations}

Throughout this paper, we consider the limiting distributions and the convergence rates of the  out-of-sample prediction risk when $n,p \rarrow \infty$ such that  $p/n = c_n \rarrow c \in (0, \infty).$
If $c>1$, the sample size $n$ is smaller than the number of parameters $p$, 
we call this case  \emph{``overparametrized"}. Otherwise when $c<1$, we call it \emph{``underparameterized"}.

As follows are some notations used in this paper.
The $p\times p$ identity matrix is denoted by $\mI_p.$
For a symmetric matrix $\mA \in \sR^{p \times p}$, we define its empirical spectral
distribution as 
\benrr
F^\mA(x) = \frac{1}{p} \sum_{i=1}^p \mathds{1}\{\lambda_i(\mA)\leq x\}
\eenrr
where $\mathds{1}\{\cdot\}$ is the indicator function, and $\lambda_i(\mA)$, $i = 1,2, \ldots p$ are the eigenvalues of $\mA.$
What's more, the notation $\xrightarrow{d}$ stands for the convergence in distribution.
Throughout this paper, 
$Z_{\alpha/2}$ is the $\alpha/2$ upper quantile of the standard normal distribution, $\lambda_{\max}(\mA)$ and $\lambda_{\min}(\mA)$ denote the largest and smallest eigenvalues of $\mA$ respectively.

In the following, we will derive  confidence intervals for both risk under various combinations of model assumptions for $c$, $\rmX$ and $\mBeta.$ Here we list all the assumptions needed in different scenarios:
\begin{itemize}
	\item[(A)] $\rvx_j\sim P_\rvx$ is of the form $\rvx_j= \mSigma^{1/2}\rvz_j$, where $\rvz_j$ is a $p$-length random vector with i.i.d. entries that have zero mean, unit variance, and a finite $4$-th order moment $\E (\rvz_{ij}^4) = \nu_4$, $i=1,\cdots,p$, $j=1,\cdots,n.$
	\item[(B1)] $\mSigma$ is a deterministic positive definite matrix, such that $\lambda(\mSigma) \geq c_0 > 0$, for all $n$, $p$ and a constant $c_0$. As $ p \rarrow \infty$, we assume that the empirical spectral distribution $F^{\mSigma}$ converges weakly to a measure $H.$	
	\item[(B2)] $\mSigma$ is an identity matrix, $\mSigma=\mI_p.$ 
 	\item[(C1)] $\mBeta$ is a nonrandom constant vector, and $\|\mBeta\|_2^2= \mBeta^\T \mBeta = r^2.$
 	\item[(C2)]  $\mBeta\sim P_\mBeta$ is independent of $\rmX$ and follows multivariate Gaussian distribution $N_p(0, \frac{r^2}{p}\mI_p)$.
\end{itemize}

\subsection{Underparametrized asymptotics}\label{sec42}

In this section, we focus on the risk of the min-norm estimator (\ref{eq12}) in the underparametrized regime.
According to Theorem~1 of \cite{hastie2019surprises}, both
$B_{\rmX,\mBeta}(\hat\mBeta,\mBeta)$ and $B_{\rmX}(\hat\mBeta,\mBeta)$ converge to $\sigma^2c/(1-c)$ almost surely.
The following theorems show that both $B_{\rmX}(\hat\mBeta,\mBeta)$ and $B_{\rmX,\mBeta}(\hat\mBeta,\mBeta)$ converge to $\sigma^2c/(1-c)$ at the rate of $1/p.$ 
Furthermore, the limiting distributions are derived by making use of the CLT for linear spectral statistics of large-dimensional sample covariance matrices.

\begin{theorem}\label{thm:underparaRX}
Suppose that the training data is generated from the model~{\rm (\ref{eq11})}, and the assumptions {\rm (A)} and {\rm (B1)} hold.
Then the first type of out-of-sample prediction risk $R_{\rmX}(\hat\mBeta,\mBeta)$ of the min-norm estimator {\rm (\ref{eq12})} satisfies that, as $n, p \rarrow \infty$ such that $p/n = c_n\rarrow c < 1$,  
\ben\label{clt1}
p\Big(R_{\rmX}(\hat\mBeta, \mBeta) - \frac{c_n \sigma^2}{1-c_n}\Big) \xrightarrow{d} N( \mu_{c}, \sigma^2_c), 
\een
where 
\benn
\mu_c=\frac{c^2\sigma^2}{(c-1)^2}+\frac{\sigma^2c^2(\nu_4-3)}{1-c} \quad \text{and} \quad \sigma_c^2=\frac{2c^3\sigma^4}{(c-1)^4}+\frac{ c^3\sigma^4(\nu_4-3)}{(1-c)^2}.
\eenn
Conclusively,
\ben\label{ci1}
P(L_{\alpha,c} \leq R_{\rmX}(\hat\mBeta, \mBeta) \leq U_{\alpha,c} ) \rarrow 1 - \alpha,
\een
where $1-\alpha$ is the confidence level and
\benrr
L_{\alpha,c}= \frac{c_n \sigma^2}{1-c_n} + \frac{1}{p}(\mu_c - Z_{\alpha/2}\sigma_c), \quad
U_{\alpha,c} = \frac{c_n \sigma^2}{1-c_n} + \frac{1}{p}(\mu_c + Z_{\alpha/2}\sigma_c).
\eenrr
\end{theorem}

Under the assumptions of Theorem~\ref{thm:underparaRX}, we know that $\mPi = \mI_p -  \hat\mSigma^{+}\hat\mSigma={\bm 0}$ and
\benn
B_{\rmX}(\hat\mBeta,\mBeta) = B_{\rmX,\mBeta}(\hat\mBeta,\mBeta) = 0, \quad V_\rmX (\hat\mBeta, \mBeta) = V_{\rmX, \mBeta} (\hat\mBeta, \mBeta) = \frac{\sigma^2}{n} \Tr(\hat\mSigma^{+}\mSigma).
\eenn 
Thus $R_{\rmX}(\hat\mBeta,\mBeta)$  equals to $R_{\rmX,\mBeta}(\hat\mBeta,\mBeta)$ and the two risk share the same asymptotic limit.

\begin{theorem}\label{thm:underparaRXB}
Under the assumptions of Theorem~\ref{thm:underparaRX}, 
the second type of out-of-sample prediction risk  $R_{\rmX, \mBeta}(\hat\mBeta, \mBeta)$ of the min-norm estimator {\rm (\ref{eq12})} satisfies that, as $n, p \rarrow \infty$ such that $p/n = c_n\rarrow c < 1$,  
\benn
p\big(R_{\rmX, \mBeta}(\hat\mBeta, \mBeta) - \frac{c_n \sigma^2}{1-c_n}\big) \xrightarrow{d} N( \mu_{c}, \sigma^2_c),
\eenn
and 
\benn
P(L_{\alpha,c} \leq R_{\rmX, \mBeta}(\hat\mBeta, \mBeta) \leq U_{\alpha,c} ) \rarrow 1 - \alpha,
\eenn
where $\mu_{c}$, $\sigma^2_c$, $L_{\alpha,c}$ and $U_{\alpha,c}$ are the same as those in Theorem~\ref{thm:underparaRX}.
\end{theorem}

\subsection{Overparametrized asymptotics}\label{sec43}
	
In this section, we consider the min-norm estimator (\ref{eq12}) in the overparametrized case.  The bias term , either $B_\rmX (\hat\mBeta, \mBeta)$ or $B_{\rmX,\mBeta}(\hat\mBeta, \mBeta)$,
is generally nonzero when $c >1$.
According to Lemma~2 of \cite{hastie2019surprises}, both $B_\rmX (\hat\mBeta, \mBeta)$ and $B_{\rmX,\mBeta}(\hat\mBeta, \mBeta)$ converge to $r^2(1-1/c)$ as $n,p \rarrow +\infty$ and $p/n \rarrow c>1.$
This implies that the bias term can influence the asymptotic behavior of the prediction risk, including the convergence rate.
Hence in order to derive the CLT of the out-of-sample prediction risk, we need to consider both the bias and variance terms in (\ref{eq13}). 

In the following, we investigate the asymptotic properties of the two prediction risk $R_\rmX(\hat\mBeta, \mBeta)$ and $R_{\rmX,\mBeta}(\hat\mBeta, \mBeta)$ under various combinations of the assumptions (A1), (B2) for $\rmX$ and scenarios (C1), (C2) for $\mBeta.$
We start with the case when $\mBeta$ is a constant vector.


\begin{theorem}\label{thm:overparaFixBeta}
Suppose that the training data is generated from the model~(\ref{eq11}), and the assumptions {\rm (A)}, {\rm (B2)} and {\rm (C1)} hold.
Then the first type of out-of-sample prediction risk, $R_{\rmX}(\hat\mBeta,\mBeta),$ of the min-norm estimator {\rm (\ref{eq12})} satisfies that,  as $n, p \rarrow \infty$ such that $p/n = c_n\rarrow c > 1$,
\ben\label{clt2}
\sqrt{p}\Big\{ R_{\rmX}(\hat\mBeta,\mBeta) - (1-\frac{1}{c_n})r^2-\frac{\sigma^2}{c_n-1} \Big\}\xrightarrow{d} N(\mu_{c,1}, \sigma_{c,1}^2),
\een
where $\mu_{c,1}=0$ and $\sigma_{c,1}^2 = \frac{2(c-1)}{c^2} r^4.$
A more practical version is to replace $\mu_{c,1}$ and $\sigma_{c,1}^2$ with
\benrr
\tilde{\mu}_{c,1} &=& \frac{1}{\sqrt{p}}\Big\{\frac{c\sigma^2}{(1-c)^2}+\frac{\sigma^2(\nu_4-3)}{c-1}\Big\},\\
\tilde{\sigma}_{c,1}^2 &=& \frac{2(c-1)}{c^2} r^4+\frac{1}{p}\Big\{ \frac{2c^3\sigma^4}{(1-c)^4}+\frac{c\sigma^4(\nu_4-3)}{(c-1)^2}\Big\}.
\eenrr
Conclusively,
\ben\label{ci2}
P(L_{\alpha,c} \leq R_{\rmX}(\hat\mBeta, \mBeta) \leq U_{\alpha,c} ) \rarrow 1 - \alpha,
\een
where $1-\alpha$ is the confidence level and
\benrr
L_{\alpha,c} &=&  (1-\frac{1}{c_n}) r^2+ \frac{\sigma^2}{c_n-1}+ \frac{1}{\sqrt{p}}(\tilde{\mu}_{c,1}- Z_{\alpha/2}\tilde\sigma_{c,1}), \\
U_{\alpha,c} &=&  (1-\frac{1}{c_n}) r^2 + \frac{\sigma^2}{c_n-1}+ \frac{1}{\sqrt{p}}(\tilde\mu_{c,1} + Z_{\alpha/2}\tilde\sigma_{c,1}).
\eenrr
\end{theorem}

\begin{remark}
Under  assumption (C1),  $B_\rmX (\hat\mBeta, \mBeta)= B_{\rmX,\mBeta}(\hat\mBeta, \mBeta)$ and  $R_\rmX (\hat\mBeta, \mBeta)= R_{\rmX,\mBeta}(\hat\mBeta, \mBeta).$
Thus Theorem~\ref{thm:overparaFixBeta} still holds if we replace $R_\rmX (\hat\mBeta, \mBeta)$ with $R_{\rmX,\mBeta}(\hat\mBeta, \mBeta).$
\end{remark}

Next we consider the case when $\mBeta$ is a random vector that follows Assumption~(C2),  we have
 
\begin{theorem}\label{thm:overparaRX}
Suppose that the training data is generated from the model~{\rm (\ref{eq11})}, and the assumptions {\rm (A)}, {\rm (B2)} and {\rm (C2)} hold.
Then, as $n, p \rarrow \infty$ such that $p/n = c_n\rarrow c > 1$, the first type of out-of-sample prediction risk, $R_{\rmX}(\hat\mBeta,\mBeta),$ of the min-norm estimator {\rm (\ref{eq12})} satisfies,
\benn
p\Big\{R_\rmX(\hat\mBeta, \mBeta)  -(1-\frac{1}{c_n})r^2 -\frac{\sigma^2}{c_n-1} \Big\} \xrightarrow{d} N( \mu_{c,2}, \sigma^2_{c,2}), 
\eenn
where 
\benn
\mu_{c,2}=\frac{c\sigma^2}{(1-c)^2}+\frac{\sigma^2(\nu_4-3)}{c-1}\quad \text{and} \quad \sigma^2_{c,2}=\frac{2c^3\sigma^4}{(1-c)^4}+\frac{c\sigma^4(\nu_4-3)}{(c-1)^2}.
\eenn
Hence we have
\benn
P(L_{\alpha,c} \leq R_{\rmX}(\hat\mBeta, \mBeta) \leq U_{\alpha,c} ) \rarrow 1 - \alpha,
\eenn
where
\benrr
L_{\alpha,c} &=& \frac{\sigma^2}{c_n-1}+( 1-\frac{1}{c_n})r^2 + \frac{1}{p} (\mu_{c,2} - Z_{\alpha/2}\sigma_{c,2}), \\
U_{\alpha,c} &=& \frac{\sigma^2}{c_n-1}+(1-\frac{1}{c_n})r^2 + \frac{1}{p}(\mu_{c,2} + Z_{\alpha/2}\sigma_{c,2}).
\eenrr
\end{theorem}

As for $R_{\rmX,\mBeta}(\hat\mBeta,\mBeta)$, we have the following theorem.
\begin{theorem}\label{thm:overparaRXB}
Suppose that the training data is generated from the model~{\rm (\ref{eq11})}, and the assumptions {\rm (A)}, {\rm (B2)} and {\rm (C2)} hold.
Then, as $n, p \rarrow \infty$ such that $p/n = c_n\rarrow c > 1$, the second type of out-of-sample prediction risk, $R_{\rmX,\mBeta}(\hat\mBeta,\mBeta),$ of the min-norm estimator {\rm (\ref{eq12})} satisfies, 
\ben\label{clt3}
\sqrt{p}\Big\{ R_{\rmX,\mBeta}(\hat\mBeta,\mBeta) -( 1-\frac{1}{c_n})r^2 -\frac{\sigma^2}{c_n-1} \Big\}\xrightarrow{d} N(\mu_{c,3}, \sigma_{c,3}^2),
\een
where $\mu_{c,3}=0$ and $\sigma_{c,3}^2= 2(1-\frac{1}{c})r^4.$
A more practical version is to replace $\mu_{c,3}$ and $\sigma_{c,3}^2$ with
\benrr
\tilde{\mu}_{c,3} &=& \frac{1}{\sqrt{p}}\left\{\frac{c\sigma^2}{(1-c)^2}+\frac{\sigma^2(\nu_4-3)}{c-1}\right\}, \\
\tilde{\sigma}_{c,3}^2 &=& 2(1-\frac{1}{c})r^4+\frac{1}{p}\left\{ \frac{2c^3\sigma^4}{(1-c)^4}+\frac{c\sigma^4(\nu_4-3)}{(c-1)^2}\right\},	
\eenrr
and the corresponding $(1-\alpha)$-confidence interval is given by
\ben\label{ci3}
P(L_{\alpha,c} \leq R_{\rmX, \mBeta}(\hat\mBeta, \mBeta) \leq U_{\alpha,c} ) \rarrow 1 - \alpha,
\een
with
\benrr
L_{\alpha,c} &=& \frac{\sigma^2}{c_n-1}+( 1-\frac{1}{c_n})r^2 + \frac{1}{\sqrt{p}}(\tilde{\mu}_{c,3}- Z_{\alpha/2}\tilde\sigma_{c,3}), \\
U_{\alpha,c} &=& \frac{\sigma^2}{c_n-1}+( 1-\frac{1}{c_n})r^2 + \frac{1}{\sqrt{p}}(\tilde\mu_{c,3} + Z_{\alpha/2}\tilde\sigma_{c,3}).
\eenrr
\end{theorem}

\begin{remark}
	If we compare the results in Theorem~\ref{thm:overparaFixBeta} and \ref{thm:overparaRXB}, we will find out that $R_{\rmX}$ with constant $\beta$ and $R_{\rmX,\mBeta}$ with random $\beta$ share the same first order limit and second order error rate $O(p^{-1/2})$. In fact, this is quite intuitive because both risk treat $\beta$ as a constant. Their differences are reflected in their  limiting variances. Nevertheless, it's very interesting to observe from Theorem~\ref{thm:overparaRX} that, $R_{\rmX}$ with random $\beta$ under the overparametrized case has smaller second order error rate $O(p^{-1})$. It enjoys the same rate {as} the underparametrized case in Theorem~\ref{thm:underparaRX}. A possible explanation would be that averaging over random $\beta$ can partially offset the curse of dimensionality, so that $R_{\rmX}$ achieves the same error rate for all $p,n$ combinations.
\end{remark}
\begin{remark}
It's worth mentioning that the only assumption regarding data distribution is Assumption (A), where only finite fourth order moment is required. Non-Gaussianity allows our theoretical results more widely applied.
\end{remark}

\section{Experiments}\label{sec:simu}

In this section, we carry out simulation experiments to examine the central limit theorems and the corresponding confidence intervals in Theorem~\ref{thm:underparaRXB} and Theorem~\ref{thm:overparaRXB}.
We generate data points from the linear model  (\ref{eq11}) and directly compute the prediction risk via the bias-variance decomposition in (\ref{eq13}).
The generative distribution $P_{\rvx}$ is taken to be the standard normal distribution.
The noise distribution $P_{\rvepsilon}$ is taken to be $N(0,1).$
In the following, we present the gap between the finite-sample distribution of the prediction risk and the corresponding limiting distribution to check the central limit theorems, and use the cover rate to measure the effectiveness of the confidence intervals.
More simulation results are relegated to the Appendix due to space limitations.

\begin{figure}[ht]
\begin{center}
\includegraphics[width=\columnwidth]{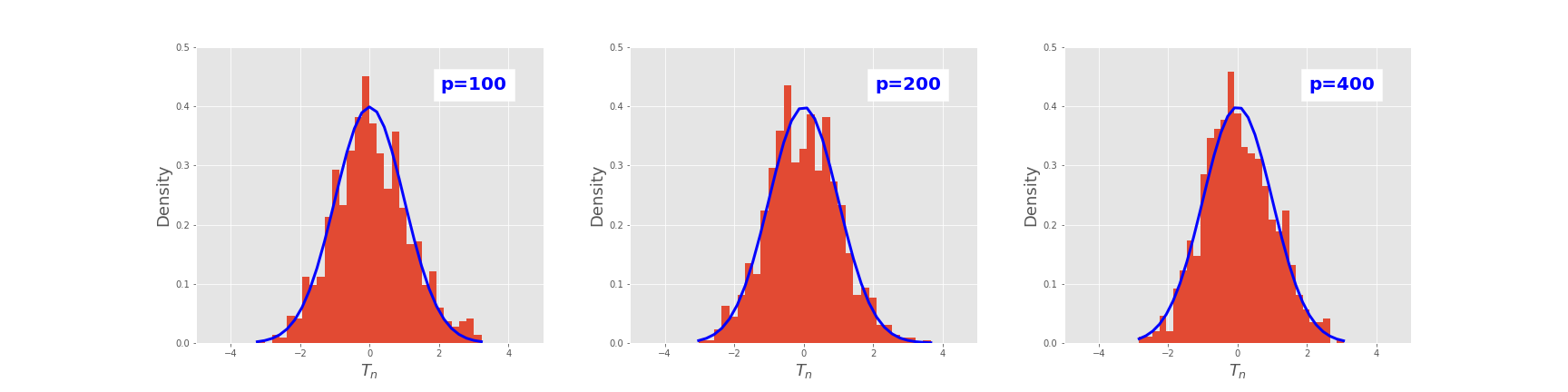}
\end{center}
\caption{The histogram of $T_n.$ The solid line is the density of the standard normal distribution.}
\label{Fig2}
\end{figure}

{\bf Example~1}. This example examines results in Theorem~\ref{thm:underparaRXB}. 
We define a statistic
\benn
T_n = \frac{p}{\sigma_c}\Big(R_{\rmX}(\hat\mBeta, \mBeta) - \sigma^2\frac{c_n}{1-c_n}\Big)-\frac{\mu_c}{\sigma_c}.
\eenn
According to Theorem~\ref{thm:underparaRXB}, 
$T_n$ weakly converges to the standard normal distribution as $n,p\rarrow \infty.$
In this example, $c=2/3$ and $p=100, 200, 400.$
The finite-sample distribution of $T_n$ is presented by the histogram of $T_n$  in Figure~\ref{Fig2} with 1000 repetitions, where the solid blue curve stands for standard normal density function.
It can be seen that the finite-sample distribution of $T_n$ is very consistent with the standard normal distribution, especially when $n,p$ become larger.
When $\alpha=0.05$, the empirical cover rates of the $95\%$-confidence interval are $93.1\%$, $93.9\%$ and $95.2\%$ for $p=100$, $200$ and $400$ respectively. All these experiments verify the correctness of our theoretical results.

\begin{figure}[ht]
\begin{center}
\includegraphics[width=\columnwidth]{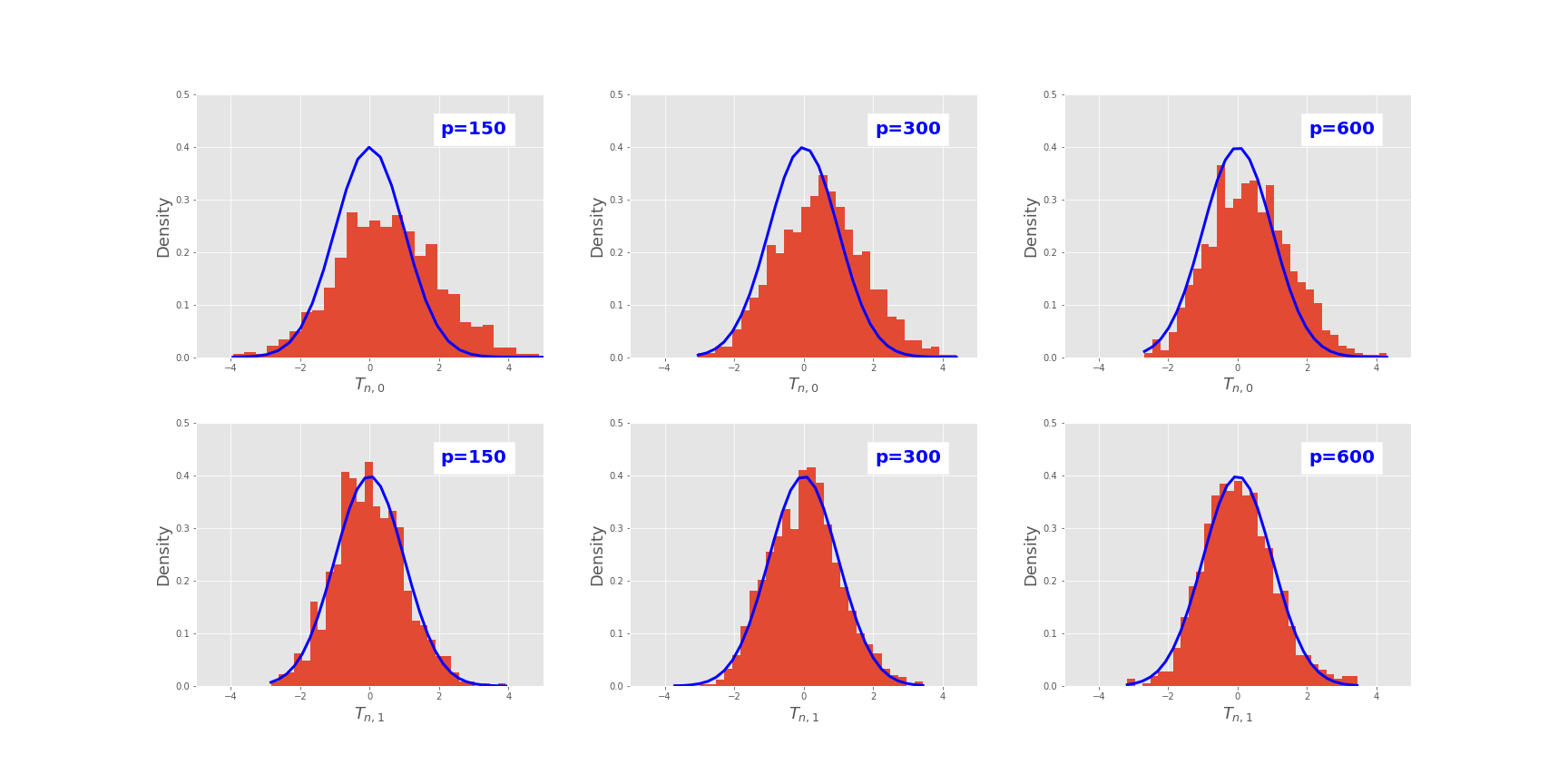}
\end{center}
\caption{The histogram of $T_{n,0}$ and $T_{n,1}.$ The solid line is the density of the standard normal distribution.} 
\label{Fig3}
\end{figure}

{\bf Example~2}. This example verifies the results in Theorem~\ref{thm:overparaRXB}. Here we define two statistics:
\benrr
T_{n,0} &=& \frac{\sqrt{p}}{\sigma_{c,3}}\Big\{ R_{\rmX}(\hat\mBeta,\mBeta) - (1-\frac{1}{c_n})r^2-\frac{\sigma^2}{c_n-1} \Big\}-\frac{\mu_{c,3}}{\sigma_{c,3}}, \\
T_{n,1} &=& \frac{\sqrt{p}}{\tilde{\sigma}_{c,3}}\Big\{ R_{\rmX}(\hat\mBeta,\mBeta) - (1-\frac{1}{c_n})r^2-\frac{\sigma^2}{c_n-1} \Big\}-\frac{\tilde{\mu}_{c,3}}{\tilde{\sigma}_{c,3}}.
\eenrr
According to Theorem~\ref{thm:overparaRXB}, both $T_{n,0}$ and $T_{n,1}$ weakly converge to the standard normal distribution as $n,p\rarrow +\infty.$
We take $c=3/2$ and $p=150, 300, 450.$ Similarly the finite-sample distributions of $T_{n,0}$ and $T_{n,1}$ are presented by the histogram of $T_{n,0}$ and $T_{n,1}$ with 1000 repetitions.
The comparison between these two statistics is shown in Figure~\ref{Fig3}. It can also be seen that the finite sample distributions of $T_{n,0}$ and $T_{n,1}$ both match the standard normal distribution quite well.
The empirical cover rates of the $95\%$-confidence interval (\ref{ci3}) are $93.8\%$, $94.7\%$ and $94.4\%$ for $p=150$, $300$ and $600$ respectively,  which further shows the validity of our theoretical results.

\bibliographystyle{iclr2021_conference}
\bibliography{reference}

\appendix
\section{Proof of theorem~\ref{thm:underparaRX} and theorem~\ref{thm:underparaRXB}}

Let $\rmX = \rmZ \mSigma^{1/2}.$ 
According to the Bai-Yin theorem (\cite{bai2008limit}), the smallest eigenvalue of $\rmZ^\T \rmZ/n$ is almost surely larger than $(1-\sqrt{c})^2/2$ for sufficiently large $n.$
Thus
\benn
\lambda_{min}(\frac{1}{n}\rmX^\T \rmX) \geq c_0 \lambda_{\min}(\frac{1}{n} \rmZ^\T \rmZ)  \geq \frac{c_0}{2}(1-\sqrt{c})^2, 
\eenn
which implies that the matrix $\rmX^\T \rmX / n$ is almost surely invertible for large $n.$ 
By Section~\ref{sec32}, $\mPi = {\bf 0}$, $B_\rmX(\hat\mBeta, \mBeta)=B_{\rmX, \mBeta}(\hat\mBeta, \mBeta)=0$ and $V_\rmX(\hat\mBeta, \mBeta) = V_{\rmX, \mBeta}(\hat\mBeta, \mBeta).$
Thus the CLT of $R_\rmX(\hat\mBeta, \mBeta)$ is same to that of $R_{\rmX, \mBeta}(\hat\mBeta, \mBeta).$
For simplicity, we focus on $R_\rmX(\hat\mBeta, \mBeta)$ in the following.
Notice that
\benrr
V_\rmX(\hat\mBeta, \mBeta) &=& \frac{\sigma^2}{n} \Tr(\hat{\mSigma}^{-1}\mSigma) \\
&=& \frac{\sigma^2}{n} \Tr\Big( \mSigma^{-1/2} \big(\frac{\rmZ^\T \rmZ}{n} \big)^{-1} \mSigma^{-1/2} \mSigma\Big) \\
&=& \frac{\sigma^2}{n} \sum_{i=1}^p \frac{1}{s_i}= \frac{\sigma^2 p}{n} \int \frac{1}{s} d F_{\rmZ}(s)
\eenrr
where $F_{\rmZ}$ is the spectral measure of $\rmZ^\T \rmZ/n.$ 
According to Theorem~1 of \cite{hastie2019surprises}, as $n, p\rarrow\infty$ such that $p/n=c_n\rightarrow c\in(0,\infty)$, $F_\rmZ(x)$ weakly converges to the standard Marcenko-Pastur law $F_c(x)$ and
\benrr
V_\rmX (\hat\mBeta,\mBeta) \rarrow \sigma^2 c \int \frac{1}{s} dF_c (s) = \sigma^2 \frac{c}{1-c}.
\eenrr
Here the standard Marcenko-Pastur law $F_c(x)$ has a density function
\benn
p_c(x)=\left\{
\begin{array}{ll}
    \frac{1}{2\pi c x}\sqrt{(b-x)(x-a)}, & \mbox{ if } a\leq x\leq b,  \\
    0, & \mbox{o.w.},
\end{array}
\right.
\eenn
where $a=(1-\sqrt{c})^2$, $b=(1+\sqrt{c})^2$ and $p_c(x)$ has a point mass $1-\frac{1}{c}$ at the origin if $c>1$.    		
Hence
\benrr
R_\rmX(\hat\mBeta, \mBeta) - \sigma^2\frac{c_n}{1-c_n} 
&=& \frac{\sigma^2 p}{n} \int \frac{1}{s} dF_{Z}(s) - \sigma^2 c_n \int \frac{1}{s} dF_{c_n} (s)\\
&=& \sigma^2 c_n \int \frac{1}{s} \big(dF_{Z}(s)- dF_{c_n} (s)\big).
\eenrr
According to Theorem~1.1 of \cite{bai2004clt}, 
\ben\label{eq:A1}
p\Big(R_\rmX(\hat\mBeta, \mBeta) - \sigma^2\frac{c_n}{1-c_n}\Big) \xrightarrow{d} N( \mu_{c}, \sigma^2_c), 
\een
where
\benr
\mu_c &=& -\frac{\sigma^2 c}{2\pi i} \oint_\gamma \frac{1}{z}\frac{c \underline{m}(z)^3 (1+\underline{m}(z))^{-3}}{\left\{1- c \underline{m}(z)^2 (1+\underline{m}(z))^{-2}\right\}^2} dz \label{eq:mu}\\
&& -\frac{\sigma^2c(\nu_4-3)}{2\pi i}\oint_\gamma \frac{1}{z}\frac{c \underline{m}(z)^3 (1+\underline{m}(z))^{-3}}{1- c \underline{m}(z)^2 (1+\underline{m}(z))^{-2}} dz,  \nonumber \\ 
\sigma^2_c &=& -\frac{\sigma^4 c^2}{2\pi^2} \oint_{\tecr{\mathcal{C}_1}}\oint_{\tecr{\mathcal{C}_2}} \frac{1}{z_1 z_2}\frac{1}{(\underline{m}(z_1) 
 - \underline{m}(z_2))^2} \frac{d}{dz_1} \underline{m}(z_1) \frac{d}{dz_2} \underline{m}(z_2) d z_1 d z_2 \label{eq:sig}\\  
&& -\frac{\sigma^4 c^3(\nu_4-3)}{4\pi^2} \oint_{\tecr{\mathcal{C}_1}}\oint_{\tecr{\mathcal{C}_2}}  \frac{1}{z_1 z_2}\frac{1}{(1+\underline{m}(z_1))^2 (1+\underline{m}(z_2) )^2}d \underline{m}(z_1)d \underline{m}(z_2). \nonumber
\eenr
Here the contours in (\ref{eq:mu}) and (\ref{eq:sig}) are closed and taken in the positive direction in the complex plane, enclosing the support of $F_\rmZ$, i.e. $[(1-\sqrt{c})^2,(1+\sqrt{c})^2]$. The Stieltjes transform $\underline{m}(z)$ satisfies the equation
\benn
z=-\frac{1}{\underline{m}}+\frac{c}{1+\underline{m}}.
\eenn
To further simplify the integrations in $\mu_c$ and $\sigma_c$,  let $z= 1+\sqrt{c}(r\xi+\frac{1}{r\xi})+c$
and perform change of variables, then we have
\benn
\underline{m}(z)=-\frac{1}{1+\sqrt{c}r\xi},\quad
dz=\sqrt{c}(r-\frac{1}{r\xi^2})d\xi,\quad d\underline{m} =\frac{\sqrt{c} r}{(1+\sqrt{c} r\xi)^2} d\xi
\eenn
and when $\xi$ moves along the unit circle $|\xi|=1$ on the complex plane, $z$ will orbit around the center point $1+c$ along an ellipse which enclosing the support of $F_\rmZ$. 
Thus
\benrr
\mu_c &=& -\frac{\sigma^2 c}{2\pi i} \oint_\gamma \frac{1}{z}\frac{c \underline{m}(z)^3 (1+\underline{m}(z))^{-3}}{(1- c \underline{m}(z)^2 (1+\underline{m}(z))^{-2})^2} dz \\
&& -\frac{\sigma^2c(\nu_4-3)}{2\pi i}\oint_\gamma \frac{1}{z}\frac{c \underline{m}(z)^3 (1+\underline{m}(z))^{-3}}{1- c \underline{m}(z)^2 (1+\underline{m}(z))^{-2}} dz\\
&=& \frac{\sigma^2 c}{2\pi i}\oint_{|\xi|=1}\frac{1}{r(\sqrt{c}+r\xi)(1+\sqrt{c}r\xi)(\xi-\frac{1}{r})(\xi+\frac{1}{r})}d\xi\\
&& +\frac{\sigma^2c(\nu_4-3)}{2\pi i} \oint_{|\xi|=1}\frac{1}{r\xi^2(\sqrt{c}+r\xi)(1+\sqrt{c}r\xi)}d\xi\\
&=&\frac{\sigma^2c^2}{(c-1)^2}+\frac{\sigma^2c^2(\nu_4-3)}{1-c}.
\eenrr
As for $\sigma_c^2$, note that
\benrr
&&\frac{1}{2\pi i}\oint_{\gamma_1}\frac{1}{z_1(\underline{m}_1-\underline{m}_2)^2}d \underline{m}_1\\
&=&\frac{1}{2\pi i}\oint_{|\xi_1|=1}\frac{1}{1+\sqrt{c}(r_1\xi_1+\frac{1}{r_1\xi_1})+c}\cdot\frac{\sqrt{c}~r_1}{(\underline{m}_2+\frac{1}{1+\sqrt{c}r_1\xi_1})^2 (1+\sqrt{c}r_1\xi_1)^2}d\xi_1\\
&=&\frac{1}{2\pi i}\oint_{|\xi_1|=1}\frac{\sqrt{c}~r_1\xi_1}{(\xi_1+\frac{\sqrt{c}}{r_1}) (r_1\xi_1\sqrt{c}+1)\left\{ (r_1\xi_1\sqrt{c}+1)\underline{m}_2+1\right\}^2}d\xi_1\\
&=&\frac{c}{(c-1)\left\{ (c-1)\underline{m}_2-1\right\}^2},
\eenrr
therefore
\benrr
&& -\frac{\sigma^4 c^2}{2\pi^2} \oiint \frac{1}{z_1 z_2 (\underline{m}_1 - \underline{m}_2)^2}d\underline{m_1}d\underline{m}_2\\
&=& \frac{2\sigma^4 c^2}{2\pi i}\oint_{|\xi_2|=1}\frac{c}{z_2(c-1)\left\{ (c-1)\underline{m}_2-1\right\}^2}d \underline{m_2}\\
&=& \frac{2\sigma^4 c^2}{2\pi i}\oint_{|\xi_2|=1}\frac{\sqrt{c}~r_2^2\xi_2}{(c-1)(1+\sqrt{c}~r_2\xi_2) (\sqrt{c}+r_2\xi_2)^3}d\xi_2=\frac{2c^3\sigma^4}{(c-1)^4}.
\eenrr
Meanwhile, 
\benrr
&&\frac{1}{2\pi i}\oint_{\gamma_1}\frac{1}{z_1}\frac{1}{(1+\underline{m}(z_1))^2} d\underline{m}(z_1)\\
&=&\frac{1}{2\pi i}\oint_{|\xi|=1}\frac{1}{\sqrt{c}\xi(1+\sqrt{c}r\xi)(\sqrt{c}+r\xi)}d\xi=\frac{1}{c-1},    
\eenrr
hence
\benn
-\frac{\sigma^4 c^3(\nu_4-3)}{4\pi^2} \oint_{\mathcal{C}_1}\oint_{\mathcal{C}_2}  \frac{1}{z_1z_2}\frac{1}{(1+\underline{m}(z_1))^2 (1+\underline{m}(z_2))^2}d \underline{m}(z_1)d \underline{m}(z_2)
=\frac{\sigma^4 c^3(\nu_4-3)}{(1-c)^2},
\eenn
and
\benn
\sigma_c^2=\frac{2c^3\sigma^4}{(c-1)^4}+\frac{\sigma^4 c^3(\nu_4-3)}{(1-c)^2}.
\eenn
Let
\benn
T_n =  \frac{p}{\sigma_c} \Big(R_\rmX(\hat\mBeta, \mBeta) - \sigma^2\frac{c_n}{1-c_n} - \frac{\mu_{c}}{p}\Big).  
\eenn
According to (\ref{eq:A1}), we have
\benn
P(L_{\alpha,c} \leq R_{\rmX, \mBeta}(\hat\mBeta, \mBeta) \leq U_{\alpha,c} ) = P(-Z_{\alpha/2} \leq T_n \leq Z_{\alpha/2}) \rarrow 1 - \alpha,
\eenn
where
\benrr
L_{\alpha,c} &=& \sigma^2\frac{c_n}{1-c_n} + \frac{1}{p}(\mu_c - Z_{\alpha/2}\sigma_c), \\
U_{\alpha,c} &=& \sigma^2\frac{c_n}{1-c_n} + \frac{1}{p}(\mu_c + Z_{\alpha/2}\sigma_c).
\eenrr

\bbox

\section{Proof of theorem~\ref{thm:overparaFixBeta}}

Notice that
\benrr
B_\rmX(\hat\mBeta,\mBeta)&=&  \mBeta^\T (\mI_p -\hat{\mSigma}^{+}\hat{\mSigma})\mBeta\\
&=&\lim_{z\rarrow 0^{+}} \mBeta^\T \big(\mI_p - (\hat\mSigma + z \mI_p )^{-1}\hat\mSigma \big)\mBeta\\
&=&\lim_{z\rarrow 0^{+}}  z \mBeta^\T(\hat\mSigma + z\mI_p)^{-1}\mBeta.
\eenrr
Since $\mBeta$ is a constant vector, we can make use of the results in Theorem 3 in \cite{Bai2007} and Theorem 1.3 in \cite{pan2008} regarding eigenvectors. 
Their works investigate the sample covariance matrix  $\rmA_p=\mT_p^{1/2} \rmX_p^\T \rmX_p \mT_p^{1/2}/n$, where $\mT_p$ is an $p\times p$ nonnegative definite Hermitian matrix with a square root $\mT_p^{1/2}$ and $\rmX_p$ is an $n \times p$ matrix with i.i.d. entries $(\rx_{ij})_{n\times p}$. 
Let $\mU_p \mLambda_p \mU_p^\T$ denote the spectral decomposition of $\rmA_p$ where $\mLambda_p=\mbox{diag}(\lambda_1,\cdots, \lambda_p)$ and $\mU_p$ is a unitary matrix consisting of the orthonormal eigenvectors of $\rmA_p$. 
Assume that $\vx_p$ is an arbitrary nonrandom unit vector and $\vy=(y_1,y_2,\cdots,y_p)^\T=\mU_p^\T \vx_p$, two empirical distribution functions based on eigenvectors and eigenvalues are defined as
\benn
F_1^{\rmA_p}(x)=\sum_{i=1}^p |y_i|^2\mathds{1}(\lambda_i\leq x), \quad F^{\rmA_p}(x)=\frac{1}{p}\sum_{i=1}^p\mathds{1}(\lambda_i\leq x).
\eenn
Then for a bounded continuous function $g(x)$, we have
\benn
\sum_{j=1}^p|\ry_j|^2 g(\lambda_j)-\frac{1}{p}\sum_{j=1}^p g(\lambda_j)=\int g(x) d F_1^{\mA_p}(x)-\int g(x)d F^{\mA_p}(x).
\eenn
The results in \cite{Bai2007} and \cite{pan2008} show that
\begin{lemma}\label{prop:fixBeta}
{\bf (Theorem 3 \cite{Bai2007} and Theorem 1.3 \cite{pan2008})} Suppose that
\begin{itemize}
	\item[(1)] $\rx_{ij}$'s are i.i.d. satisfying $\E (\rx_{ij})=0$, $\E(|\rx_{ij}|^2)=1$ and $\E (|\rx_{ij}|^4)<\infty$;
	\item[(2)]$\vx_p\in \sC^p$, $\|\vx_p\|=1$, $\lim_{n, p\rarrow\infty}p/n=c\in(0,\infty)$;
	\item[(3)]$\mT_p$ is nonrandom Hermitian non-negative definite with with its spectral norm bounded in $p$, with $H_p=F^{\mT_p}\xrightarrow{d} H$ a proper distribution function and $\vx_p^\T (\mT_p-z\mI_p)^{-1}\vx_p\rightarrow m_{F^H}(z)$, where $m_{F^H}(z)$ denotes the Stieltjes transform of $H(t)$;
	\item[(4)] $g_1,\cdots, g_k$ are analytic functions on an open region of the complex plain which contains the real interval 
	\benn
	\Big[ \liminf_p\lambda_{min}(\mT_p) \mathds{1}_{(0,1)}(c)(1-\sqrt{c})^2, \,\, \limsup_p\lambda_{max}(\mT_p) \mathds{1}_{(0,1)}(c)(1+\sqrt{c})^2\Big];
	\eenn
	\item[(5)] as $n, p\rarrow\infty$,
	\benn
	\sup_z \sqrt{n} \Big\| \vx_p^\T \big(\underline{m}_{F^{c_n,H_p}}(z)\mT_p-\mI_p\big)^{-1} \vx_p-\int\frac{1}{1+t\underline{m}_{F^{c_n,H_p}}(z)}d H_n(t) \Big\|\rarrow 0.
	\eenn
\end{itemize}
Define $G_p(x)=\sqrt{n} (F_1^{\mA_p}(x) - F^{\mA_p}(x))$, then the random vectors
\benn
\left(\int g_1(x)d G_p(x), \cdots, \int g_k(x)d G_p(x)\right)
\eenn
forms a tight sequence and converges weakly to a Gaussian vector $\rx_{g_1},\cdots, \rx_{g_k}$ with mean zero and covariance function
\benn
\Cov(\rx_{g_1},\rx_{g_2})=-\frac{1}{2\pi^2}\int_{\sC_1}\int_{\sC_2}g_1(z_1)g_2(z_2)\frac{(z_2\underline{m}_2-z_1\underline{m}_1)^2}{c^2z_1z_2(z_2-z_1)(\underline{m}_2-\underline{m}_1)}d z_1 d z_2.
\eenn
The contours $\sC_1$, $\sC_2$ are disjoint, both contained in the analytic region for the functions $(g_1,\cdots, g_k)$ and enclose the support of $F^{c_n,H_p}$ for all large $p$.
\begin{itemize}
	\item[(6)]   If $H(x)$ satisfies
	\[
	\int\frac{d H(t)}{(1+t\underline{m}(z_1))(1+t\underline{m}(z_2))}=\int\frac{d H(t)}{1+t\underline{m}(z_1)}\int\frac{d H(t)}{1+t\underline{m}(z_2)},
	\]
\end{itemize}
then the covariance function can be further simplified to
\[
\Cov(\rx_{g_1}, \rx_{g_2})=\frac{2}{c} (\int g_1(x)g_2(x)d F^{c,H}(x)-\int g_1(x) d F^{c,H}(x)\int g_2(x) d F^{c,H}(x) ).
\]
\end{lemma}

Recall that $B_\rmX(\hat \mBeta,\mBeta) = \lim_{z \rarrow 0^+} z\mBeta^\T(\hat\mSigma+z\mI_p)^{-1}\mBeta$. Let $g(x)=1/(x+z)$ and $\vx_p = \mBeta/r$. Then we have 
\benn
\int g(x) dG_n(x)=\sqrt{n} \Big( \frac{1}{r^2}\mBeta^\T(\hat\mSigma+z\mI_p)^{-1}\mBeta - \int g(x) dF_{c_n}(x)\Big),
\eenn
where $F_{c_n}(x)$ is the standard Marcenko-Pastur law. It is not difficult to check that under Assumptions (A1), (B1) and (C1),  all the conditions \emph{(1)-(6)} in Lemma~\ref{prop:fixBeta} are satisfied.

To proceed further, denote $a=(1-\sqrt{c})^2$, $b=(1+\sqrt{c})^2$. 
If $c$ is replaced by $c_n$, $a$ and $b$ are denoted by $a_n$ and $b_n$ respectively.
By some algebraic calculations, we have 
\benrr
\int g(x) d F_{c_n}(x)&=&(1-\frac{1}{c_n})\cdot\frac{1}{z}+\int_{a_n}^{b_n}\frac{1}{x+z}\cdot\frac{1}{2\pi c_n x}\sqrt{(b_n-x)(x-a_n)} d x\\
&=&(1-\frac{1}{c_n})\cdot\frac{1}{z}-\frac{-1+c_n+z-\sqrt{c_n^2+2c_n(z-1)+(1+z)^2}}{2c_nz},
\eenrr
and
\benrr
\Var(\rx_g) &=& \frac{2}{c} \left( \int \{g(x)\}^2d F_c(x)-\big\{\int g(x) dF_c(x)\big\}^2 \right)\\
&=& \frac{2}{c}\Big( (1-\frac{1}{c})\cdot\frac{1}{z^2}+\int_{a}^{b}\frac{1}{(x+z)^2}\cdot\frac{1}{2\pi c x}\sqrt{(b-x)(x-a)} d x \Big)\\
&& -\frac{2}{c}\Big( (1-\frac{1}{c})\cdot\frac{1}{z}+\int_{a}^{b}\frac{1}{x+z}\cdot\frac{1}{2\pi c x}\sqrt{(b-x)(x-a)} d x\Big)^2.
\eenrr
Therefore,
\benn
\lim_{z\rarrow 0^+} z \int g(x) d F_{c_n}(x)=1-\frac{1}{c_n} \quad \text{and} \quad
\lim_{z\rarrow 0^+} z^2  \Var(\rx_g)=\frac{2(c-1)}{c^3}.
\eenn
Furthermore, as $n,p\rarrow \infty$, $p/n=c_n\rarrow c>1$, 
\benn
\sqrt{n} \Big( B_\rmX(\hat\mBeta,\mBeta) -(1-\frac{1}{c_n})r^2 \Big) \xrightarrow{d}  N\Big( 0, \frac{2(c-1)}{c^3} r^4\Big).
\eenn
This can be rewritten as
\benn
\sqrt{p} \Big( B_\rmX(\hat\mBeta,\mBeta) -(1-\frac{1}{c_n})r^2 \Big) \xrightarrow{d}  N\Big( 0, \frac{2(c-1)}{c^2} r^4\Big).
\eenn
Next we deal with the variance term $V_{\rmX}(\hat\mBeta, \mBeta)$.
According to the Assumption \emph{(B1)}, the variance term is 
\benrr
V_\rmX (\hat\mBeta, \mBeta) = \frac{\sigma^2}{n} \Tr\{\hat\mSigma^{+}\} = \frac{\sigma^2}{n} \sum_{i=1}^n \frac{1}{s_i},
\eenrr
where $s_i$, $i=1,\dots, n$ are the nonzero eigenvalues of $\rmX^\T \rmX/n.$
Let $\{t_i,~i = 1, \ldots n\}$ denote the non-zero eigenvalues of $\rmX \rmX^\T/p$, then we have 
\benrr
V_\rmX (\hat\mBeta, \mBeta) = \frac{\sigma^2}{p} \sum_{i=1}^n \frac{1}{t_i} = \frac{\sigma^2 n}{p} \int \frac{1}{t} dF_{\rmX \rmX^\T/p}(t) \rarrow \frac{\sigma^2}{c-1}.
\eenrr
By interchanging the role of $p$ and $n$, from the result in Theorem~\ref{thm:underparaRX}, as $n, p \rarrow \infty$,  $p/n=c_n\rightarrow c > 1$,  we have,
\benn
\sum_{i=1}^n\frac{1}{t_i}-\frac{n}{1-c'_n}\xrightarrow{d} N\Big(\frac{c'}{(c'-1)^2}+\frac{c'(\nu_4-3)}{1-c'},~\frac{2c'}{(c'-1)^4}+\frac{c'(\nu_4-3)}{(1-c')^2} \Big).
\eenn
where $c'_n=n/p=1/c_n$, $c'=1/c$.
This result can be rewritten as
\benn
\sum_{i=1}^n\frac{1}{t_i}-\frac{p}{c_n-1}\xrightarrow{d} N \Big( \frac{c}{(1-c)^2}+\frac{(\nu_4-3)}{c-1},~\frac{2c^3}{(1-c)^4}+\frac{c(\nu_4-3)}{(c-1)^2} \Big).
\eenn
Hence the CLT of $V_{\rmX}(\hat\mBeta, \mBeta)$ is given by
\benn
p\Big( V_\rmX (\hat\mBeta, \mBeta)-\frac{\sigma^2}{c_n-1} \Big) \xrightarrow{d} N\Big(\frac{c\sigma^2}{(1-c)^2}+\frac{\sigma^2(\nu_4-3)}{c-1},~\frac{2c^3\sigma^4}{(1-c)^4}+\frac{c\sigma^4(\nu_4-3)}{(c-1)^2}\Big).
\eenn
Notice that 
$
\Cov\Big( B_{\rmX}(\hat\mBeta,\mBeta),~V_{\rmX}(\hat\mBeta, \mBeta)\Big)=0. 
$
According to the consistency rate and the limiting distribution of $B_{\rmX}(\hat\mBeta, \mBeta)$ and $V_{\rmX}(\hat\mBeta, \mBeta)$,
we know that the bias $B_{\rmX}(\hat\mBeta,\mBeta)$ is the leading term of $R_{\rmX}(\hat\mBeta,\mBeta)$.
This implies that
\benn
\sqrt{p}\Big\{ R_{\rmX}(\hat\mBeta,\mBeta) - ( 1-\frac{1}{c_n} ) \|\mBeta\|_2^2-\frac{\sigma^2}{c_n-1} \Big\}\xrightarrow{d} N \big( 0, \sigma_{c,1}^2 \big),
\eenn
where $\sigma_{c,1}^2 = 2(c-1)r^4/c^2.$
A practical version of this CLT is given by
\benn
\sqrt{p}\Big\{ R_{\rmX}(\hat\mBeta,\mBeta) - ( 1-\frac{1}{c_n} ) \|\mBeta\|_2^2-\frac{\sigma^2}{c_n-1} \Big\}\xrightarrow{d} N \big( \tilde{\mu}_{c,1}, \tilde{\sigma}_{c,1}^2 \big),
\eenn
where 
\benrr
\tilde{\mu}_{c,1} &=& \frac{1}{\sqrt{p}}\Big\{\frac{c\sigma^2}{(1-c)^2}+\frac{\sigma^2(\nu_4-3)}{c-1}\Big\},\\
\tilde{\sigma}_{c,1}^2 &=& \frac{2(c-1)}{c^2} r^4+\frac{1}{p}\Big\{ \frac{2c^3\sigma^4}{(1-c)^4}+\frac{c\sigma^4(\nu_4-3)}{(c-1)^2}\Big\}.
\eenrr

\section{Proof of theorem~\ref{thm:overparaRX}}

First we consider the bias term $B_\rmX (\hat\mBeta, \mBeta)$.		
By Assumption (A1), (B1), and (C2),
\benrr
B_\rmX (\hat\mBeta, \mBeta) &=& \E[\mBeta^\T \Pi \mSigma \Pi \mBeta| \rmX] =\E[\mBeta^\T\Pi\mBeta |\rmX] \\
&=& \Tr\left\{ (\mI_p - \hat\mSigma^{+}\hat\mSigma) \E(\mBeta\mBeta^\T|\rmX)\right\}\\
&=& \frac{r^2}{p}\Tr\{\mI_p - {\hat\mSigma}^{+}{\hat\mSigma}\} = r^2(1-n/p).
\eenrr
Alternatively, we can rewrite the bias as
\benrr
B_\rmX (\hat\mBeta, \mBeta) &=& \lim_{z \rarrow 0^{+}} \E[\mBeta^\T (\mI_p - (\hat\mSigma + z \mI_p)^{-1}\hat\mSigma\big) \mBeta |\rmX ]\\
&=& \lim_{z \rarrow 0^{+}} \E[ z \mBeta^\T  (\hat\mSigma + z \mI_p)^{-1} \mBeta |\rmX ]\\
&=&  \lim_{z \rarrow 0^{+}} z \frac{r^2}{p}  \Tr(\hat\mSigma + z \mI_p)^{-1}.
\eenrr	
Define that $f_n(z) =  z \frac{r^2}{p}  \Tr(\hat\mSigma + z \mI_p)^{-1}.$ 
Notice that $|f_n(z)|$ and $|f'_n(z)|$ are bounded above.
By the Arzela-Ascoli theorem, we deduce that $f_n(z)$ converges uniformly to its limit.
Under Assumption \emph{(C2)}, by the Moore-Osgood theorem, almost surely,
\benrr
\lim_{n,p\rarrow \infty} B_\rmX (\hat\mBeta, \mBeta) &=&  \lim_{z \rarrow 0^{+}} \lim_{n,p\rarrow \infty} z \frac{r^2}{p}  \Tr(\hat\mSigma + z \mI_p)^{-1}  \\
&=&  \lim_{z \rarrow 0^{+}} \lim_{n,p\rarrow \infty} z\frac{r^2}{p}\Tr\left( \frac{1}{n}\rmX\rmX^{\T}+z\mI_n \right)^{-1},\\
\eenrr
In fact, 
\benn
\lim_{n,p\rarrow \infty} B_\rmX (\hat\mBeta, \mBeta)=r^2\lim_{z \rarrow 0^{+}}\lim_{n,p\rarrow \infty} z m_n(-z),
\eenn
where $m_n(z)$ is the Stieltjes transform of empirical spectral distribution of $\hat\mSigma=\rmX^{\T}\rmX/n.$ 
According to Theorem 2.1 in \cite{zheng15} and Lemma 1.1 in \cite{bai2004clt}, the truncated version of $p(m_n(z)-m(z))$ converges weakly to a two-dimensional Gaussian process $M(\cdot)$ satisfying 
\benrr
\E[M(z)] &=& \frac{c\underline{m}^3(1+\underline{m})}{\left\{ (1+\underline{m})^2-c\underline{m}^2\right\}^2}+\frac{c(\nu_4-3)\underline{m}^3}{(1+\underline{m})\left\{(1+\underline{m})^2-c\underline{m}^2\right\}},
\eenrr
and
\benrr
\Cov\big(M(z_1),M(z_2)\big) &=& 2\Big\{ \frac{\underline{m}'(z_1)\underline{m}'(z_2)}{( \underline{m}(z_1)-\underline{m}(z_2))^2}-\frac{1}{(z_1-z_2)^2}\Big\}\\
&& +\frac{c(\nu_4-3)\underline{m}'(z_1)\underline{m}'(z_2)}{( 1+\underline{m}(z_1))^2( 1+\underline{m}(z_2))^2},
\eenrr
where $\underline{m}=\underline{m}(z)$ represents the Stieltjes transform of limiting spectral distribution of companion matrix $\rmX \rmX^\T /n$ satisfying the equation
\benn
z=-\frac{1}{\underline{m}}+\frac{c}{1+\underline{m}}, \quad \underline{m}(z)=-\frac{1-c}{z}+cm(z).
\eenn
When $p>n$, we can actually solve $\underline{m}(z)$ equation and obtain that
\benrr
\underline{m}(z) &=& \frac{-1+c-z+\sqrt{-4z+(1-c+z)^2}}{2z},\\
m(z) &=& \frac{1-c-z+\sqrt{-4z+(1-c+z)^2}}{2cz}.
\eenrr
Therefore, by some algebraic calculations, we have
\begin{align*}
\lim_{n,p\rarrow \infty}	B_\rmX (\hat\mBeta, \mBeta)=&~\lim_{n,p\rarrow \infty}r^2\lim_{z \rarrow 0^{+}} z m_n(-z)=r^2\lim_{z \rarrow 0^{+}} \left\{zm(-z)+z(1-\frac{1}{c})\frac{1}{z}\right\}\\
=&~\lim_{n,p\rarrow \infty}r^2\lim_{z \rarrow 0^{+}} z\frac{n}{p}\underline{m}_n(z)=r^2\frac{1}{c}\lim_{z \rarrow 0^{+}} z\underline{m}(-z)\\
=&~r^2( 1-\frac{1}{c}).
\end{align*}
Moreover, 
\benrr
\Var\big(M(z)\big) &=& \lim_{z_1\rightarrow z_2=z} \Cov \big(M(z_1),M(z_2) \big)\\
&=&\frac{2\underline{m}'(z)\underline{m}'''(z)-3(\underline{m}''(z))^2}{6( \underline{m}'(z))^2}+\frac{c(\nu_4-3)( \underline{m}'(z))^2}{( 1+\underline{m}(z))^4}.
\eenrr
By substituting of the explicit form of $\underline{m}(z)$, we can easily derive that
\benn
\lim_{z \rarrow 0^{+}} z\E[M(-z)]=0, \quad \lim_{z \rarrow 0^{+}}  z^2{\Var}(M(-z))=0,
\eenn
which means that the second order limit of $B_\rmX (\hat\mBeta, \mBeta)$ is still $r^2( 1-1/c)$.
All in all, $B_\rmX (\hat\mBeta, \mBeta)$ is identical with a constant $r^2( 1- 1/c)$ in distribution. 

On the other hand, by Assumption (B1), 
\benrr
V_\rmX (\hat\mBeta, \mBeta) = \frac{\sigma^2}{n} \Tr\{\hat\mSigma^{+}\} = \frac{\sigma^2}{n} \sum_{i=1}^n \frac{1}{s_i},
\eenrr
where $s_i$, $i=1,\dots, n$ are the nonzero eigenvalues of $\rmX^\T \rmX/n.$
Similar to the proof of Theorem~\ref{thm:overparaFixBeta},  the CLT of $V_{\rmX}(\hat\mBeta, \mBeta)$ is given by
\benn
p\Big( V_\rmX (\hat\mBeta, \mBeta)-\frac{\sigma^2}{c_n-1} \Big) \xrightarrow{d} N\Big(\frac{c\sigma^2}{(1-c)^2}+\frac{\sigma^2(\nu_4-3)}{c-1},~\frac{2c^3\sigma^4}{(1-c)^4}+\frac{c\sigma^4(\nu_4-3)}{(c-1)^2}\Big).
\eenn
Combining the results of $B_\rmX (\hat\mBeta, \mBeta)$ and $V_\rmX (\hat\mBeta, \mBeta)$, we have
\benn
p\Big\{R_\rmX(\hat\mBeta, \mBeta)  -r^2( 1-\frac{1}{c_n})- \frac{\sigma^2}{c_n-1} \Big\} \xrightarrow{d} N( \mu_{c,2}, \sigma_{c,2}^2),
\eenn
where
\benn
\mu_{c,2}=\frac{c\sigma^2}{(1-c)^2}+\frac{\sigma^2(\nu_4-3)}{c-1}, \quad \sigma_{c,2}^2=\frac{2c^3\sigma^4}{(1-c)^4}+\frac{c\sigma^4(\nu_4-3)}{(c-1)^2}.
\eenn

\section{Proof of theorem~\ref{thm:overparaRXB}}
		
Note that under Assumption \emph{(B1)} and \emph{(C2)}, $B_{\rmX,\mBeta} (\hat\mBeta, \mBeta)=\mBeta^\T\Pi\mBeta=\mBeta^\T(\mI_p - \hat\mSigma^{+}\hat\mSigma)\mBeta$. If we directly consider $\mBeta^\T(\mI_p - \hat\mSigma^{+}\hat\mSigma)\mBeta$, we can make use of the asymptotic results for quadratic forms Theorem 7.2 in \cite{BY08} stated as follows.
	
\begin{lemma}\label{prop:quad} {\bf (Theorem 7.2 in \cite{BY08}) } 
Let $\{ \mA_n=[a_{ij}(n)]\}$ be a sequence of $n\times n$ real symmetric matrices, $\left\{ \rvx_i\right\}_{i \in \mathbb{N}}$ be a sequence of i.i.d. $K$ dimensional real random vectors, with $\E(\rvx_i)=0$, $\E(\rvx_i \rvx_i^{\T})=(\gamma_{ij})_{K\times K}$ and $\E[\|\rvx_i\|^4] < \infty$. Denote
\benn
\rvx_i=(\rx_{\ell i})_{K\times 1},\quad \rmX(\ell)=(\rx_{\ell 1},\cdots, \rx_{\ell n})^\T, \quad \ell=1,\cdots, K, ~i=1,\cdots, n,
\eenn
assume the following limits exist
\benn
\omega=\lim_{n\rightarrow\infty}\frac{1}{n}\sum_{i=1}^n a_{ii}^2(n), \quad \theta=\lim_{n\rarrow \infty} \frac{1}{n}\Tr \mA_n^2.
\eenn
Then the $K$-dimensional random vectors
\[
\rvz_n=(\rz_{n,\ell})_{K\times 1}, \quad \rz_{n,\ell}=\frac{1}{\sqrt{n}}\big( \rmX(\ell)^\T\mA_n \rmX(\ell)-\gamma_{\ell\ell}\Tr\{\mA_n\} \big),\quad 1\leq \ell \leq K,
\]
converge weakly to a zero-mean Gaussian vector with covariance matrix $\mD=\mD_1+\mD_2$ where
\[
[\mD_1]_{\ell\ell'}=\omega\left\{\E(x_{\ell 1}^2x_{\ell' 1}^2)-\gamma_{\ell \ell}\gamma_{\ell' \ell'}\right\},~[\mD_2]_{\ell\ell'}=(\theta-\omega)(\gamma_{\ell\ell'}\gamma_{\ell'\ell}+\gamma_{\ell\ell'}^2),~ 1\leq \ell,\ell'\leq K.
\]
\end{lemma}

According to the results in Lemma~\ref{prop:quad}, let $\mA_n=\Pi=\mI_p - \hat\mSigma^{+}\hat\mSigma$, then we have, as $p\rarrow \infty$,
\benn
\sqrt{p}\Big\{ \mBeta^\T\Pi\mBeta-\frac{r^2}{p}\Tr(\Pi) \Big\} \xrightarrow{d} N( 0, d^2=d_1^2+d_2^2),
\eenn
where 
\benn
\omega = \lim_{p\rarrow \infty}\frac{1}{p}\sum_{i=1}^p\Pi_{ii}^2, \quad \theta=\lim_{p\rarrow \infty}\frac{1}{p}\Tr (\Pi^2)=1-\frac{1}{c},
\eenn
and
\benrr
d_1^2 &=& \omega\left\{ \E( x_{\ell 1}^2 x_{\ell 1}^2 )-\gamma_{\ell\ell}^2\right\}
=\omega \big(\frac{p^2}{r^4}~\E (\beta_i^4)-1 \big)r^4,\\
d_2^2 &=& (\theta-\omega)(\gamma_{\ell\ell}^2+\gamma_{\ell\ell}^2)=2(\theta-\omega)r^4.
\eenrr
Since in the proof of Theorem~\ref{thm:overparaRX}, we have already shown that 
\begin{gather*}
 	\frac{r^2}{p}\Tr(\Pi)=r^2( 1-\frac{n}{p}).
\end{gather*}
In particular, if $\mBeta$ follows multivariate Gaussian distribution, i.e. $\mBeta\sim N_p( 0, \frac{r^2}{p}\mI_p)$, then as $p\rarrow\infty$,
\[
\sqrt{p}\Big\{ B_{\rmX,\mBeta}(\hat\mBeta,\mBeta) -r^2( 1-\frac{n}{p}) \Big\}\xrightarrow{d} N \Big( 0, 2(1-\frac{1}{c})r^4\Big).
\]
Moreover, $V_{\rmX,\mBeta} (\hat\mBeta, \mBeta)=V_\rmX (\hat\mBeta, \mBeta)$,  we have already proved in Theorem~\ref{thm:overparaRX} that
\benn
p( V_{\rmX,\mBeta} (\hat\mBeta, \mBeta)-\frac{\sigma^2}{c_n-1} )\xrightarrow{d} N\Big( \frac{c\sigma^2}{(1-c)^2}+\frac{\sigma^2(\nu_4-3)}{c-1},~\frac{2c^3\sigma^4}{(1-c)^4}+\frac{c\sigma^4(\nu_4-3)}{(c-1)^2} \Big).
\eenn
Note that $\Cov( B_{\rmX,\mBeta}(\hat\mBeta,\mBeta),~V_{\rmX,\mBeta}(\hat \mBeta, \mBeta))=0.$
According to the consistency rate of $B_{\rmX,\mBeta}(\hat\mBeta, \mBeta)$ and $V_{\rmX,\mBeta}(\hat\mBeta, \mBeta)$,
we know that the bias $B_{\rmX}(\hat\mBeta,\mBeta)$ is the leading term of $R_{\rmX, \mBeta}(\hat\mBeta,\mBeta)$.
This implies that
\benn
\sqrt{p}\Big\{ R_{\rmX,\mBeta}(\hat\mBeta,\mBeta) -r^2( 1-\frac{1}{c_n})-\frac{\sigma^2}{c_n-1} \Big\}\xrightarrow{d} N( 0, \sigma_{c,3}^2),
\eenn
where $\sigma_{c,3}^2 = 2 r^4 (1-1/c).$
A practical version of this CLT is given by
\benn
\sqrt{p}\Big\{R_{\rmX,\mBeta}(\hat\mBeta,\mBeta) -r^2( 1-\frac{1}{c_n})-\frac{\sigma^2}{c_n-1} \Big\}\xrightarrow{d} N( \tilde{\mu}_{c,3}, \tilde{\sigma}_{c,3}^2),
\eenn
where 
\benrr
\tilde{\mu}_{c,3} &=& \frac{1}{\sqrt{p}}\Big\{\frac{c\sigma^2}{(1-c)^2}+\frac{\sigma^2(\nu_4-3)}{c-1}\Big\},\\
\tilde{\sigma}_{c,3}^2 &=& 2(1-\frac{1}{c})r^4+\frac{1}{p}\Big\{ \frac{2c^3\sigma^4}{(1-c)^4}+\frac{c\sigma^4(\nu_4-3)}{(c-1)^2}\Big\}.	
\eenrr

\section{More experiments}

\subsection{More results of Example~1}
\begin{figure}[htbp]
\begin{center}
\includegraphics[width=\columnwidth]{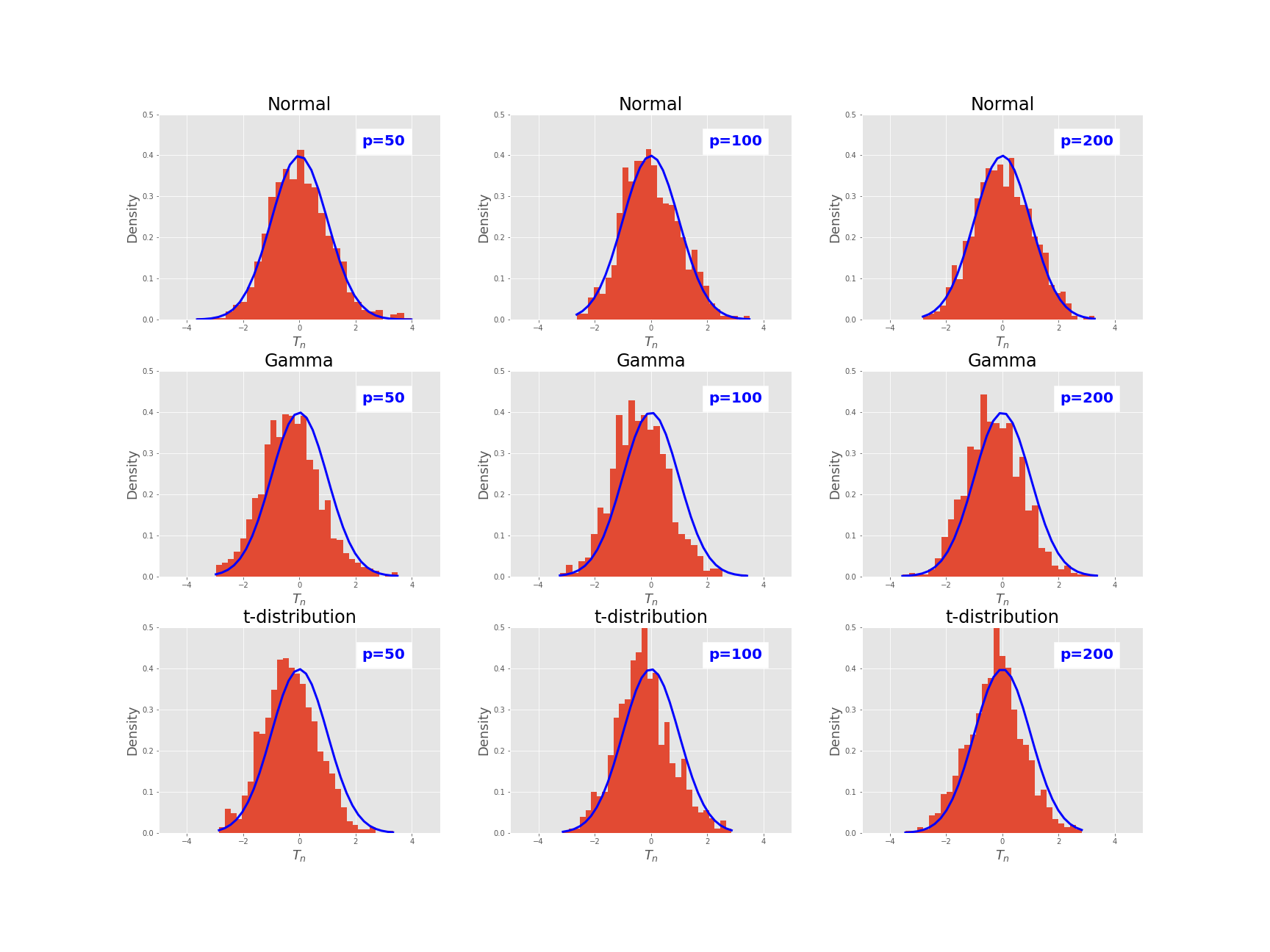}
\end{center}
\caption{The histogram of $T_n.$ The solid line is the density of the standard normal distribution.}
\label{FigA1}
\end{figure}

\begin{figure}[htbp]
\begin{center}
\includegraphics[width=\columnwidth]{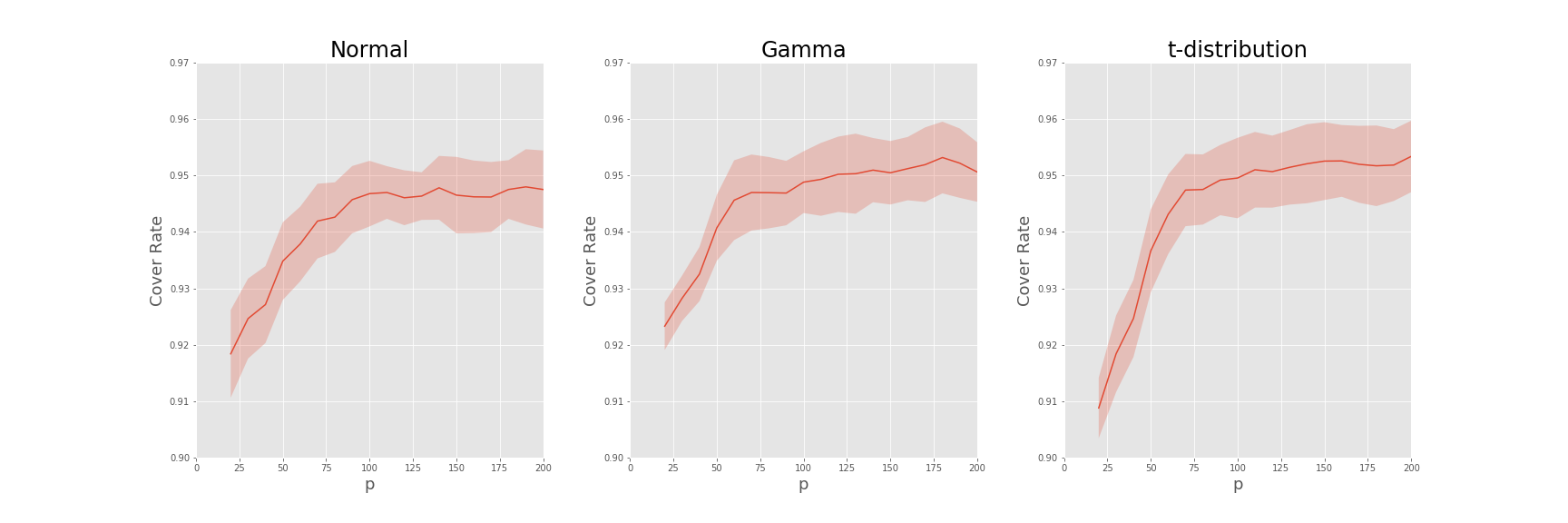}
\end{center}
\caption{The cover rate of the confidence interval (\ref{ci1}) as $p$ creases. The confidence level is $95\%.$}
\label{FigA2}
\end{figure}

This example checks Theorem~\ref{thm:underparaRXB}. 
We define a statistic
\benn
T_n = \frac{p}{\sigma_c}\Big(R_{\rmX}(\hat\mBeta, \mBeta) - \sigma^2\frac{c_n}{1-c_n}\Big)-\frac{\mu_c}{\sigma_c}.
\eenn
According to Theorem~\ref{thm:underparaRXB}, 
$T_n$ weakly converges to the standard normal distribution as $n,p\rarrow \infty.$
In this example, $c=1/2$ and $p=50, 100, 200.$
To make sure the assumption (A) holds, the generative distribution $P_{\rvx}$ is taken to be the standard normal distribution, the centered gamma with shape $4.0$ and scale $0.5$, and the normalized Student-t distribution with $6.0$ degree of freedom.
The finite-sample distribution of $T_n$ is estimated by the histogram of $T_n$ under 1000 repetitions.
The results are presented in Figure~\ref{FigA1}.
One can find that the finite-sample distribution of $T_n$ tends to the standard normal distribution as $n,p \rarrow +\infty.$
When $\alpha=0.05$, the empirical cover rates of the $95\%$-confidence interval are reported in Figure~\ref{FigA2}.

\subsection{More results of Example~2}

\begin{figure}[htbp]
\begin{center}
\includegraphics[width=\columnwidth]{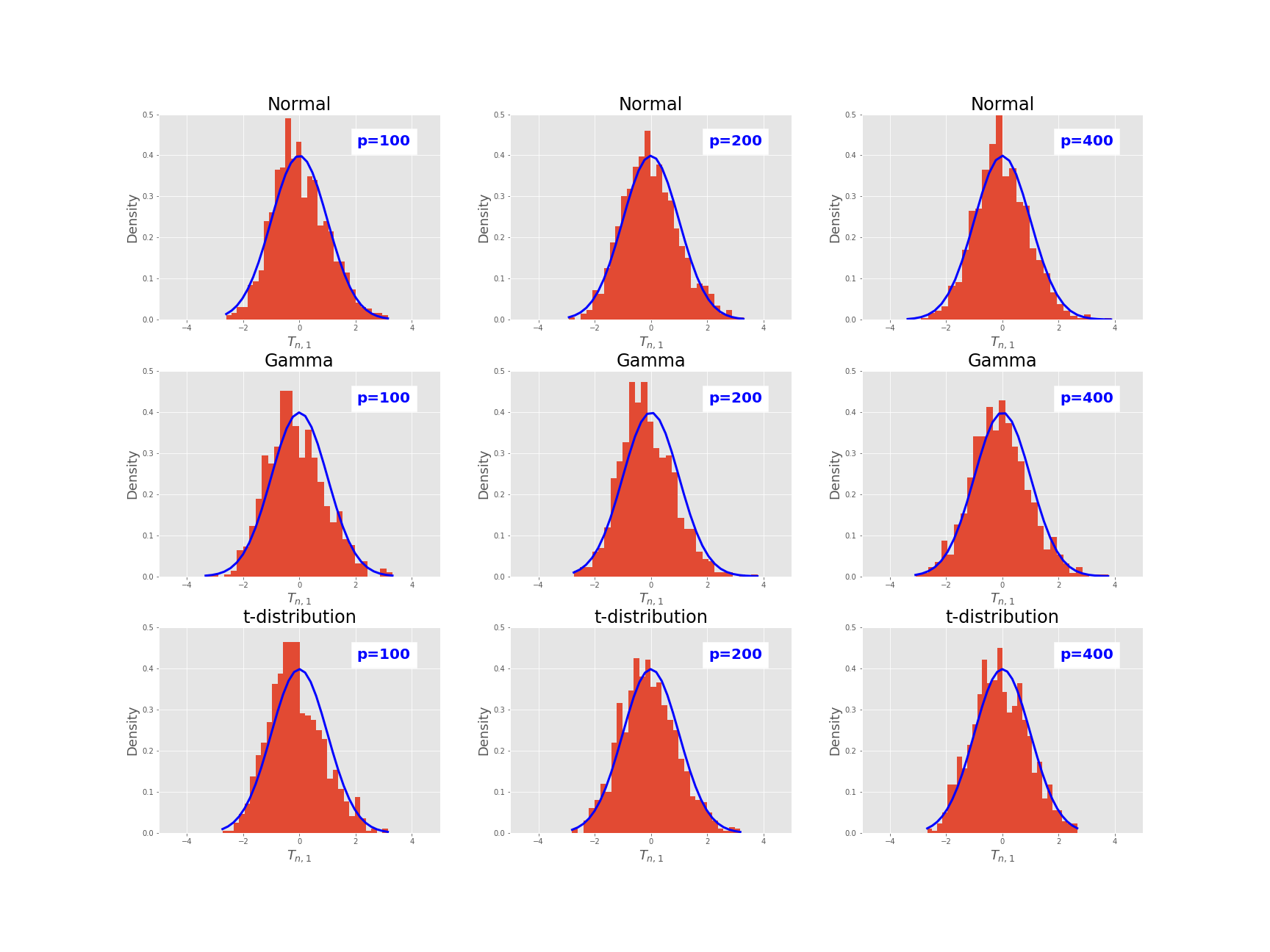}
\end{center}
\caption{The histogram of $T_{n,1}.$ The solid line is the density of the standard normal distribution.}
\label{FigA3}
\end{figure}

\begin{figure}[htbp]
\begin{center}
\includegraphics[width=\columnwidth]{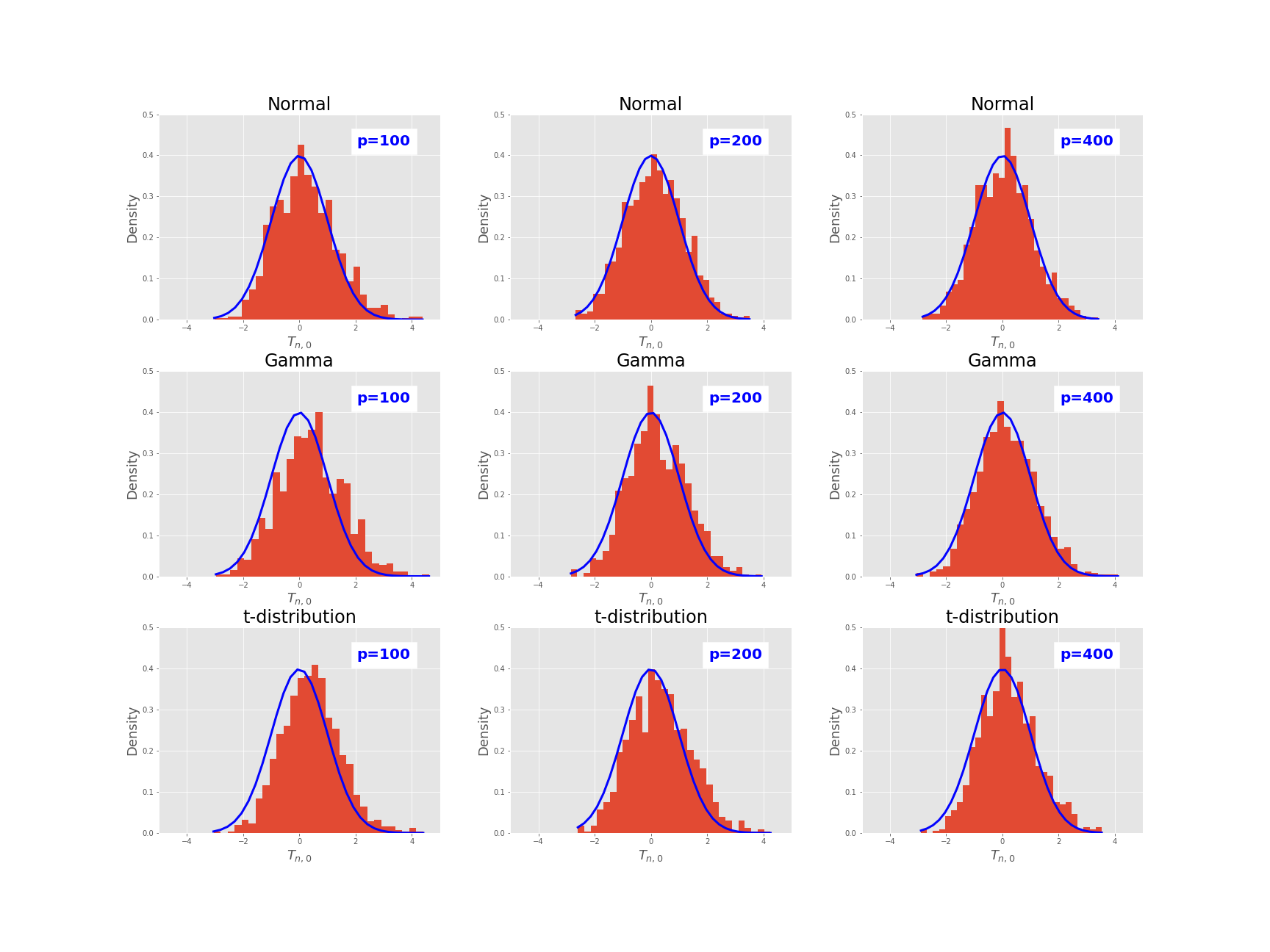}
\end{center}
\caption{The histogram of $T_{n,0}.$ The solid line is the density of the standard normal distribution.}
\label{FigA4}
\end{figure}

\begin{figure}[htbp]
\begin{center}
\includegraphics[width=\columnwidth]{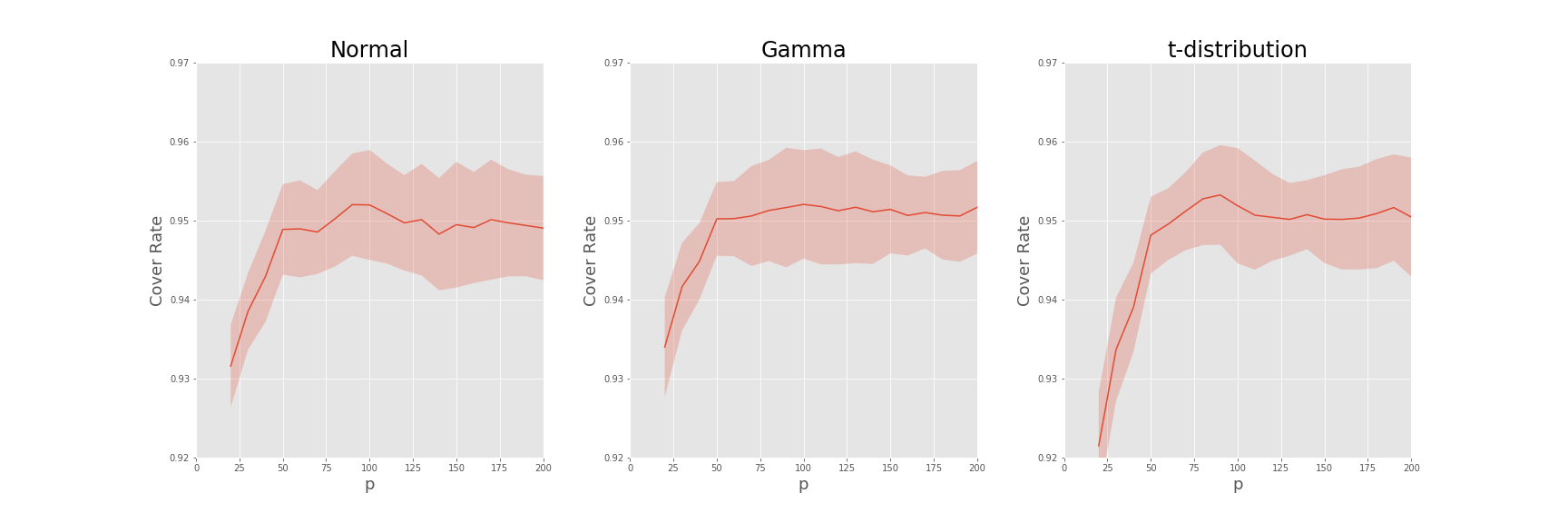}
\end{center}
\caption{The cover rate of the confidence interval (\ref{ci3}) as $p$ creases. The confidence level is $95\%.$}
\label{FigA5}
\end{figure}

The Example~2 checks Theorem~\ref{thm:overparaRXB}.
Here we consider the standardized statistics:
\benrr
T_{n,0} &=& \frac{\sqrt{p}}{\sigma_{c,3}}\Big\{ R_{\rmX}(\hat\mBeta,\mBeta) - (1-\frac{1}{c_n})r^2-\frac{\sigma^2}{c_n-1} \Big\}-\frac{\mu_{c,3}}{\sigma_{c,3}}, \\
T_{n,1} &=& \frac{\sqrt{p}}{\tilde{\sigma}_{c,3}}\Big\{ R_{\rmX}(\hat\mBeta,\mBeta) - (1-\frac{1}{c_n})r^2-\frac{\sigma^2}{c_n-1} \Big\}-\frac{\tilde{\mu}_{c,3}}{\tilde{\sigma}_{c,3}}.
\eenrr
According to the central limit theorem (\ref{clt3}) and its practical version, both $T_{n,0}$ and $T_{n,1}$ weakly converge to the standard normal distribution as $n,p\rarrow +\infty.$
We take $c=2$ and $p=100, 200, 400.$
The finite-sample distributions of $T_{n,0}$ and $T_{n,1}$ are estimated by the histogram of $T_{n,0}$ and $T_{n,1}$ under 1000 repetitions.
The results are presented in Figure~\ref{FigA3} and Figure~\ref{FigA4}.
When $\alpha=0.05$, the empirical cover rates of the $95\%$-confidence interval (\ref{ci3}) are reported in Figure~\ref{FigA5}.

\newpage

\subsection{Example~3}

\begin{figure}[htbp]
\begin{center}
\includegraphics[width=\columnwidth]{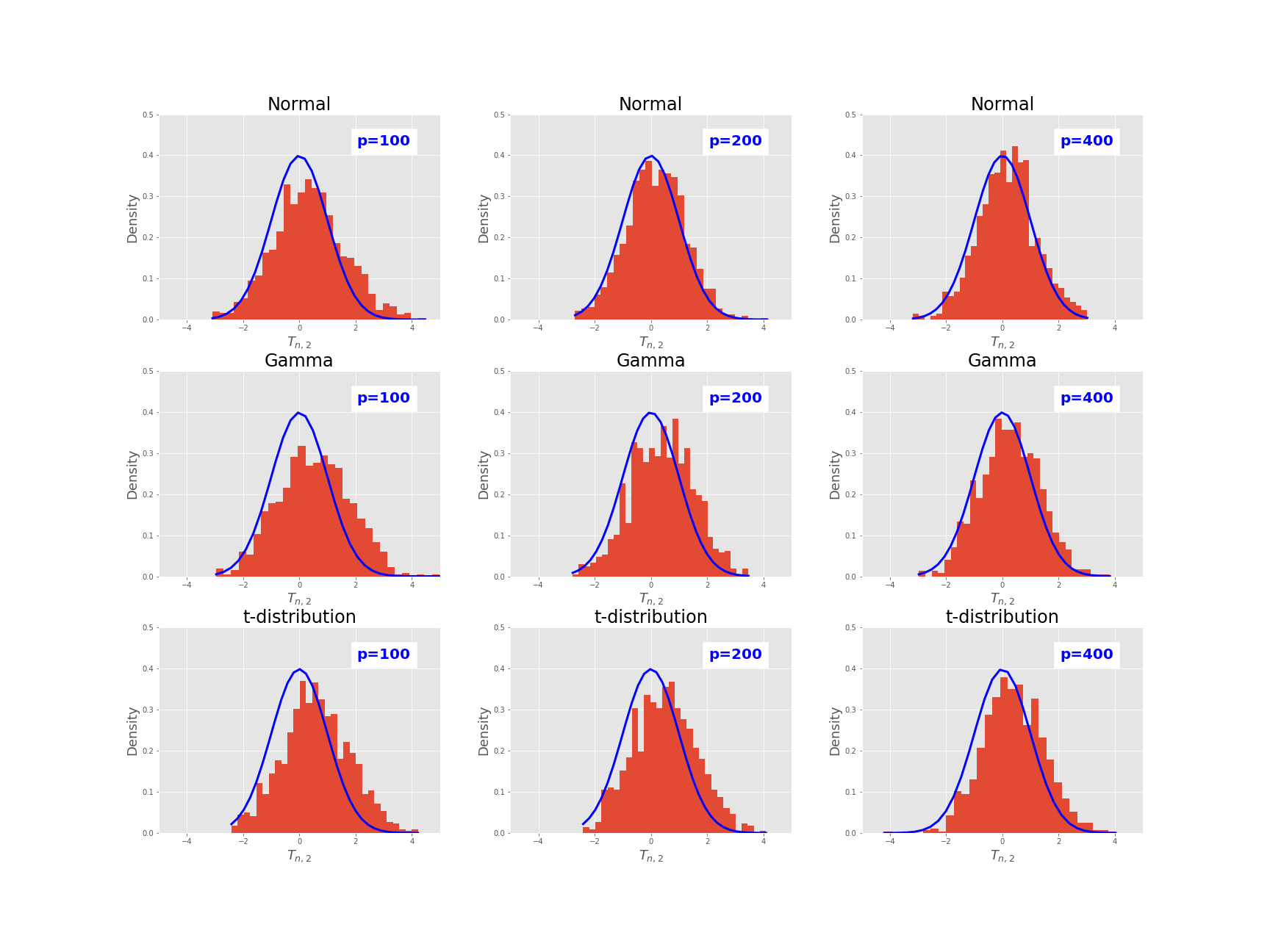}
\end{center}
\caption{The histogram of $T_{n,2}.$ The solid line is the density of the standard normal distribution.}
\label{FigA6}
\end{figure}

\begin{figure}[htbp]
\begin{center}
\includegraphics[width=\columnwidth]{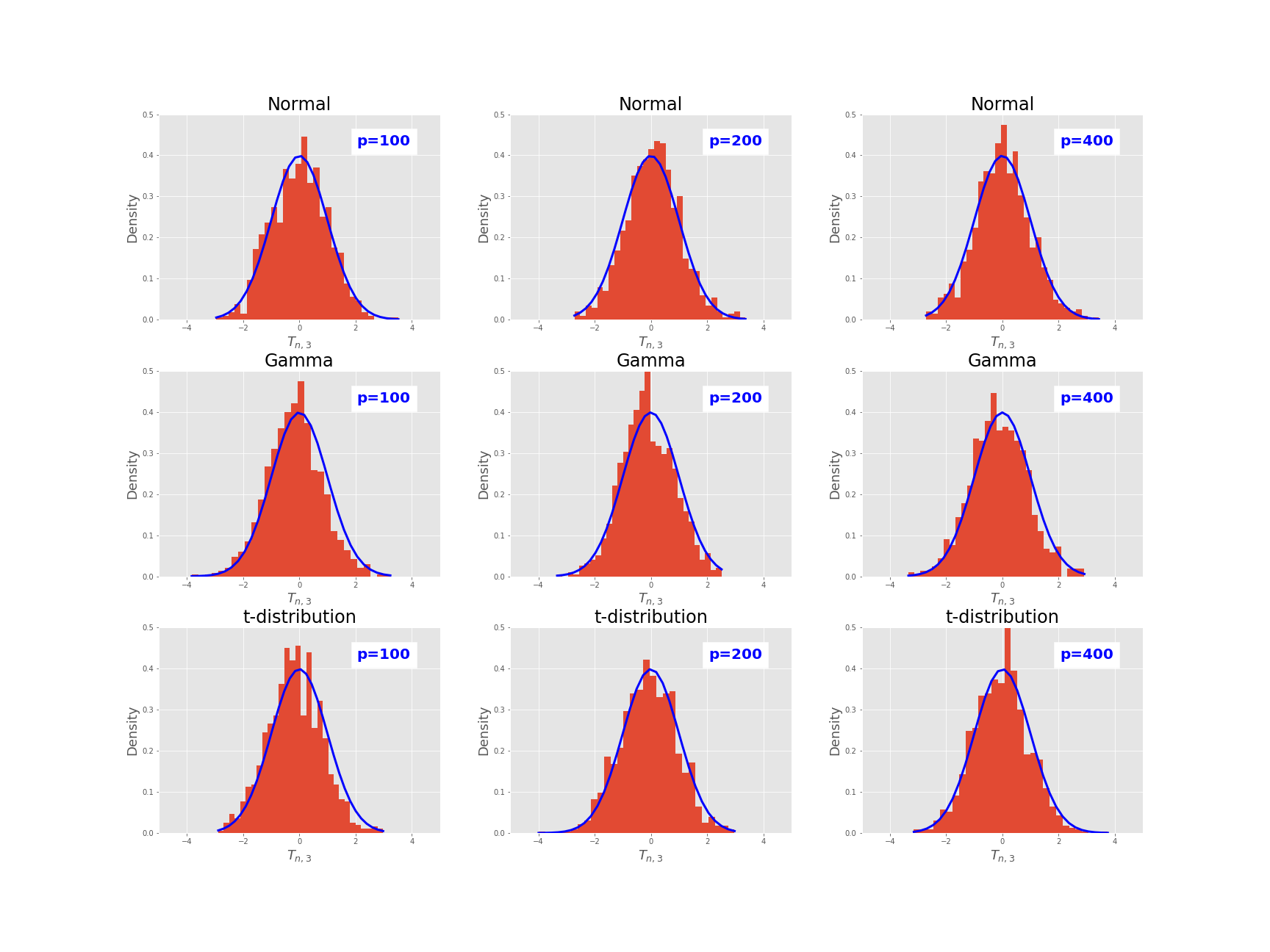}
\end{center}
\caption{The histogram of $T_{n,3}.$ The solid line is the density of the standard normal distribution.}
\label{FigA7}
\end{figure}

\begin{figure}[htbp]
\begin{center}
\includegraphics[width=\columnwidth]{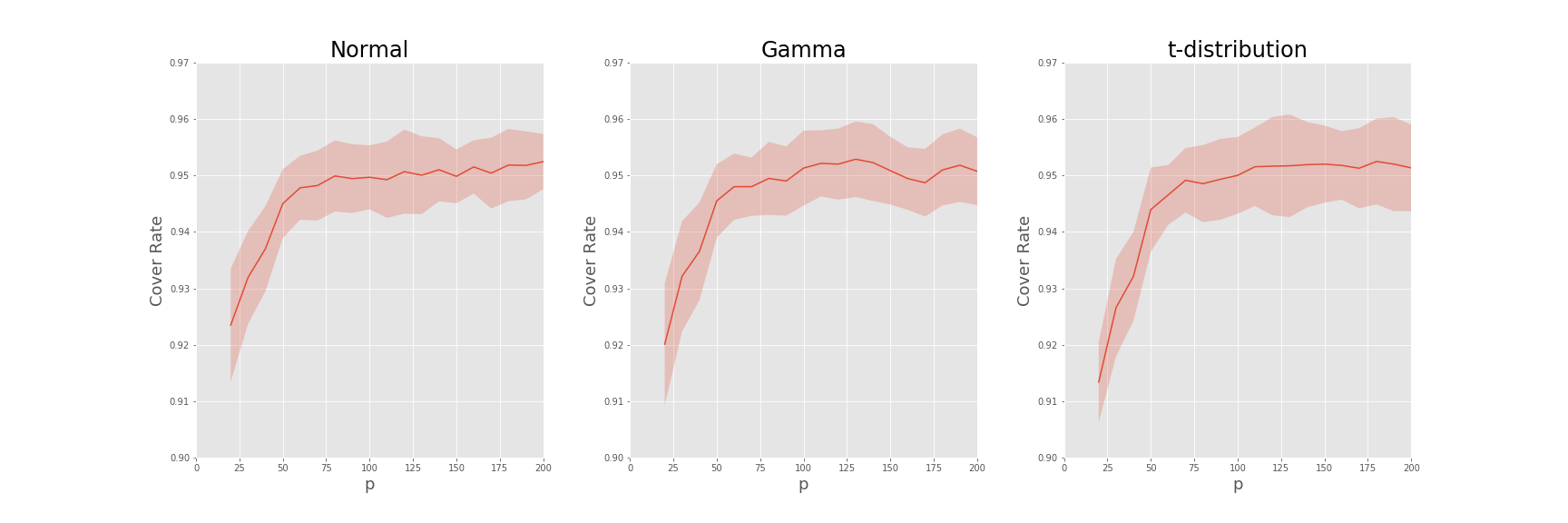}
\end{center}
\caption{The cover rate of the confidence interval (\ref{ci2}) as $p$ creases. The confidence level is $95\%.$}
\label{FigA8}
\end{figure}

This example checks Theorem~\ref{thm:overparaFixBeta}.
To proceed further, we denote two statistics:
\benrr
T_{n,2} &=& \frac{\sqrt{p}}{\sigma_{c,1}}\Big\{ R_{\rmX}(\hat\mBeta,\mBeta) - (1-\frac{1}{c_n})r^2-\frac{\sigma^2}{c_n-1} \Big\}-\frac{\mu_{c,1}}{\sigma_{c,1}}, \\
T_{n,3} &=& \frac{\sqrt{p}}{\tilde{\sigma}_{c,1}}\Big\{ R_{\rmX}(\hat\mBeta,\mBeta) - (1-\frac{1}{c_n})r^2-\frac{\sigma^2}{c_n-1} \Big\}-\frac{\tilde{\mu}_{c,1}}{\tilde{\sigma}_{c,1}}.
\eenrr
According to the central limit theorem (\ref{clt2}) and its practical version, both $T_{n,2}$ and $T_{n,3}$ weakly converge to the standard normal distribution as $n,p\rarrow +\infty.$
We take $c=2$ and $p=100, 200, 400.$
The finite-sample distributions of $T_{n,2}$ and $T_{n,3}$ are estimated by the histogram of $T_{n,2}$ and $T_{n,3}$ under 1000 repetitions.
The results are presented at Figure~\ref{FigA6} and Figure~\ref{FigA7}. 
One can see that the finite-sample distributions of $T_{n,2}$ and $T_{n,3}$ are close to the standard normal distribution, and the finite-sample performance of $T_{n,3}$ is better than that of $T_{n,2}.$ 
When $\alpha=0.05$, the empirical cover rates of the $95\%$-confidence interval (\ref{ci2}) are reported in Figure~\ref{FigA8}.

\end{document}